\documentclass[runningheads]{llncs}
\usepackage[T1]{fontenc}
\usepackage{graphicx}
\usepackage{amsmath}
\usepackage{amssymb}
\usepackage{bm}
\usepackage{booktabs}
\usepackage{multirow}
\usepackage{subfig}
\usepackage{xcolor} 
\usepackage{cite}
\usepackage{marvosym} 
\begin{document}
\title{Bridging Discrete Marks and Continuous Dynamics: Dual-Path Cross-Interaction for Marked Temporal Point Processes}
\titlerunning{Dual-Path Cross-Interaction for Marked Temporal Point Processes}

\author{Yuxiang Liu, Qiao Liu, Tong Luo, Yanglei Gan, Peng He, and Yao Liu\Letter}

\authorrunning{Y. Liu et al.}

\institute{University of Electronic Science and Technology of China, Chengdu, Sichuan, China \\
202321081212@std.uestc.edu.cn, qliu@uestc.edu.cn, tongluo1128@std.uestc.edu.cn, \\
yangleigan@std.uestc.edu.cn, hepenglk@std.uestc.edu.cn, liuyao@uestc.edu.cn}

\maketitle

\begin{abstract}
Predicting irregularly spaced event sequences with discrete marks poses significant challenges due to the complex, asynchronous dependencies embedded within continuous-time data streams. Existing sequential approaches capture dependencies among event tokens but ignore the continuous evolution between events, while Neural Ordinary Differential Equation (Neural ODE) methods model smooth dynamics yet fail to account for how event types influence future timing. To overcome these limitations, we propose NEXTPP, a dual-channel framework that unifies discrete and continuous representations via Event-granular Neural Evolution with Cross-Interaction for Marked Temporal Point Processes. Specifically, NEXTPP encodes discrete event marks via a self-attention mechanism, simultaneously evolving a latent continuous-time state using a Neural ODE. These parallel streams are then fused through a cross-attention module to enable explicit bidirectional interaction between continuous and discrete representations. The fused representations drive the conditional intensity function of the neural Hawkes process, while an iterative thinning sampler is employed to generate future events. Extensive evaluations on five real-world datasets demonstrate that NEXTPP consistently outperforms state-of-the-art models.  The source code can be found at \url{https://github.com/AONE-NLP/NEXTPP}.

\keywords{Marked Temporal Point Processes \and Neural Ordinary Differential Equation \and Event Forecasting \and Neural Hawkes Processes.}
\end{abstract}

\section{Introduction}
Event sequences consist of discretely distributed, irregularly timed occurrences whose intervals encode rich temporal dependencies absent in uniformly sampled time series \cite{du2016recurrent, upadhyay2018deep}. Such asynchronous data arise in diverse applications, from information diffusion in social networks \cite{sharma2021identifying} and patient monitoring in healthcare \cite{xu2016patient} to user behavior in e-commerce and seismic activity forecasting \cite{fox2016spatially}. Modeling these sequences requires frameworks capable of jointly handling strictly positive inter-event times and categorical “marks” (e.g., event type or location). 

Temporal point processes (TPPs) \cite{daley2003introduction} provide a unified mathematical foundation by defining a non-negative intensity function that governs the instantaneous event occurrence intensity given past observations. Classical TPPs, such as Poisson process \cite{kingman1992poisson} and self-exciting Hawkes processes \cite{hawkes1971spectra} leverage fixed, parametric intensity functions and enjoy well-established statistical properties, but their restrictive assumptions often fail to capture the nonlinear dynamics exhibited by real-world event streams.

\begin{figure}[t]
\centering
\includegraphics[width=0.66\columnwidth]{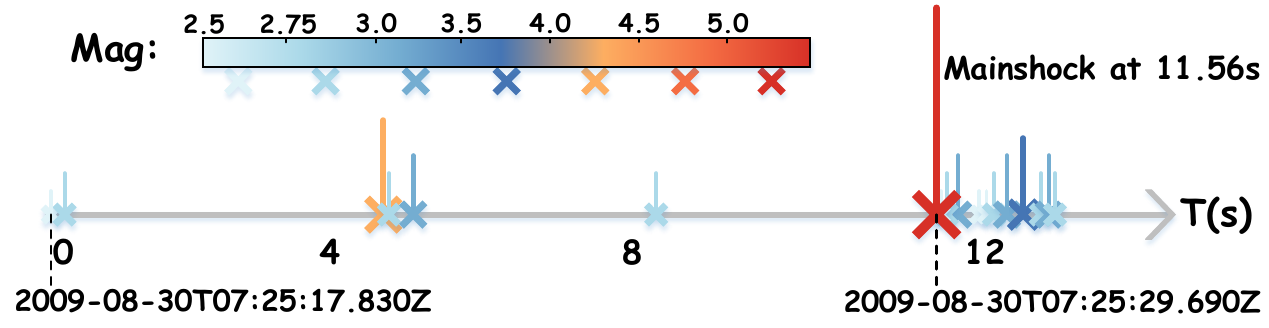}
\caption{Seismic timeline for the \textit{August 30, 2009} sequence, showing 18 events within 12.3 s. A magnitude 5.2 mainshock at 11.56s marks the transition from foreshocks to aftershocks, illustrating the classic three-phase seismic cycle.}
\label{intro}
\end{figure}

Marked TPPs (MTPPs) extend the framework by associating each event with a mark, thereby moving from univariate to multivariate modeling. Contemporary MTPP methods fall into two distinct categories: \textbf{Discrete-time models}, typically based on Recurrent Neural Networks (RNNs) or Transformer \cite{du2016recurrent, zuo2020transformer}, excel in learning dependencies among event tokens yet disregard the continuous nature of time. \textbf{Continuous-time models}, including Neural Ordinary Differential Equations and latent Stochastic Differential Equations formulations \cite{chen2018neural, rubanova2019latent}, evolve hidden representations smoothly over real time but typically do not integrate observed event marks, thereby missing explicit dependencies between marks and their temporal evolution. Consequently, neither strand fully captures the bidirectional information flow between discrete event marks and their latent continuous-time dynamics, leaving critical gaps in the current schemes.

To illustrate, consider the 18 consecutive seismic event sequence occurred in California, USA on August 30, 2009 in Figure \ref{intro}. Over just 12.3 s, small foreshocks (magnitude 2.5-4.0) occur at irregular intervals that evolve into an $\mathbf{M}$5.2 mainshock at 11.56s, immediately followed by densely clustered aftershocks of varying magnitudes. A discrete-time approach could learn the magnitude progression but would lacks modeling of temporal dependencies, whereas a continuous‑time ODE can accurately fit event timestamps but lacks modeling of correlations between different earthquakes. Thus, temporal dependencies between events in the continuous-time dimension are fundamentally coupled with mutual influences among events in the discrete-event dimension—where continuous-time dependencies provide dynamic context for discrete event interactions, while discrete event influences in turn guide the evolutionary trajectory of continuous-time dependencies. We require a unified framework to model the interplay between continuous and discrete dynamics.

To bridge this gap, we introduce NEXTPP, a dual-channel architecture that integrates Neural Evolution with Cross-Interaction for Marked Temporal Point Process. Specifically, the discrete stream encodes event representation via self-attention, while the continuous stream evolves each latent state via a Neural ODE to capture fine-grained temporal evolution between events. A cross-attention block then fuses the two streams, allowing event marks to influence future timings and temporal context to inform mark generation. This cross-interaction mechanism allows event magnitudes to adjust future timing predictions, while temporal context refines mark forecasts. The fused representations drive the conditional intensity function of the neural Hawkes process, while an iterative thinning sampler is employed to generate future events. Our main contributions are as follows:

\begin{enumerate}
    \renewcommand{\labelenumi}{\arabic{enumi})}
    \setlength{\itemsep}{0.5\baselineskip}
    \setlength{\parsep}{0pt}
    \setlength{\topsep}{0.5\baselineskip}

    \item We propose an event-granularity-based sequential evolution strategy, which achieves the modeling of complex temporal dependencies in temporal point processes while strictly preserving the global structural consistency of the Hawkes process.
    
    \item We establish the influence of historical events on the current event, enhancing the model's representational capacity for event evolution via bidirectional semantic alignment between continuous state trajectories and discrete event representations.
    
    \item Extensive experiments on five real-world datasets show that our proposed model attains more precise predictions and superior interpretability compared with state-of-the-art methods.
\end{enumerate}

\section{Preliminaries and Related Works}
\subsection{Marked Temporal Point Processes}
MTPPs are stochastic models that characterize the joint distribution of event times and their associated categorical types (marks). Formally, an event sequence of length $L$ is denoted by $\mathcal{S}=\{(t_i,m_i)\}_{i=1}^{L}$, where the $t_i\in(0,T)$ is the arrival time of $i$-th event, and $m_i \in \{1,2,...M\}$ is its corresponding type. Assuming that the event times are strictly increase monotonically, so the inter-event interval satisfies ${\bigtriangleup}t_i=t_i-t_{i-1} \in {\mathbb{R}}^+$, with the convention $\bigtriangleup t_1=t_1$. Let $\mathcal{H}_t=\{e_j=(t_j,m_j)|t_j < t\}$ denote the historical event sequence until time $t$, the event distribution of the MTPPs is then denoted as the following conditional intensity function:
\begin{equation}  \label{intensity}
    {\lambda}^{i}(t,m) \triangleq \lambda(t,m|\mathcal{H}_{t_i}) \\
    =\lim_{\bigtriangleup t \to 0}\frac{\mathbb{P}(t_i\in[t,t+\bigtriangleup t],m_i=m|\mathcal{H}_{t_i})}{\bigtriangleup t}.
\end{equation}

For notational brevity, the explicit dependence on $\mathcal{H}_{t_i}$ is absorbed into the superscript $i$ in Eq \eqref{intensity}. The conditional intensity function characterizes the instantaneous rate that the event with a mark $m$ will occur during the time interval $[t, t+\bigtriangleup t]$ under the given historical event sequence $H_{t_{i}}$. The generalized conditional probability is then expressed as:
\begin{equation}
p^i(t,m)=\lambda^i(t,m)e^{-\int^t_{t_{i-1}}\sum^M_{l=1}\lambda^i(s,l)ds}.
\end{equation}

Classical statistical point process models typically rely on parametric intensity functions. The Poisson process \cite{kingman1992poisson} is the most fundamental independent increment process. Hawkes processes \cite{hawkes1971spectra} introduce self‑excitation: each event transiently boosts future intensity via a decaying kernel, capturing clustering. Self‑correcting processes \cite{isham1979self} instead impose self‑inhibition, with each event suppressing subsequent intensity to maintain balance. These conventional approaches merely model purely excitatory or inhibitory effects, but are incapable of capturing the intricate dependency structures embedded within real-world event streams.

Deep learning–based TPP models replace handcrafted intensities with learned representations. Early approaches used RNNs to encode event histories via the iterative hidden-state updates \cite{du2016recurrent, mei2017neural}, but encountered bottlenecks in capturing long-range dependencies and computational efficiency. Attention-based paradigms for MTPPs were further established to capturing global dependencies in event sequences. The Self-attentive and Transformer Hawkes processes were individually proposed in \cite{zhang2020self} and \cite{zuo2020transformer}. The continuous-time Transformer was established in \cite{yang2022transformer} to summarize the past events. The Transfomer-based variational autoencoder was employed in \cite{zhou2022neural} to approximate the ground-truth intensity functions. 

Certainly, a parallel line of works dedicate to address the intractable integral of the conditional intensity that appears in the likelihood of density for MTPPs. The policy-based method of reinforcement learning was utilized in \cite{upadhyay2018deep} to detour the accumulation of intensity. A intensity-free alternative was proposed in \cite{rubanova2019latent} to directly parameterize the probability density function on the observed event sequences. Diffusion medels \cite{ho2020denoising, yuan2023spatio} and score matching techniques\cite{li2023smurf} are widely adopted to circumvent the costly integral in density.

\subsection{Neural Ordinary Differential Equations}
Neural Ordinary Differential Equations (Neural ODEs) cast a neural network $f$ as the time-varying vector field of an ODE, deriving the state $\mathbf{h}$ along the trajectory: $\frac{\mathrm{d}\mathbf{h}(t)}{\mathrm{d}t}=f(\mathbf{h}(t), t; \theta)$, where $\theta$ denotes the network parameters. The practical breakthrough came with the adjoint sensitivity method \cite{chen2018neural}, which delivers exact gradients at a constant-memory cost, breaking the large-memory deadlock of traditional back-propagation. In the wake of Neural ODEs and their flexible learning algorithm, a growing family of variants have rapidly emerged \cite{kidger2020neural, dupont2019augmented, song2020score}. More recently, the insights of control theory have been woven into these Neural ODE models, yielding to faster convergence and provable adversarial robustness \cite{rodriguez2022lyanet}. Departing from discrete-time approximations that struggle to capture fine-grained temporal dynamics for MTPPs, several recent studies conceptualize discrete event sequences as continuous dynamics via Neural ODEs. ODE-RNNs \cite{rubanova2019latent} define the continuous-time hidden states of RNNs via Neural ODEs to naturally handle the irregular time gaps in MTPPs. Attentive Continuous-time Normalizing Flows was proposed in \cite{chen2020neural} to learn the complex distribution of events. A stochastic process view is provided in \cite{zhang2024neural} to model the continuous diffusion process between events and instantaneous jumps at event times.

\section{Methodology}

\begin{figure*}[t]
\centering
\includegraphics[width=0.94\textwidth]{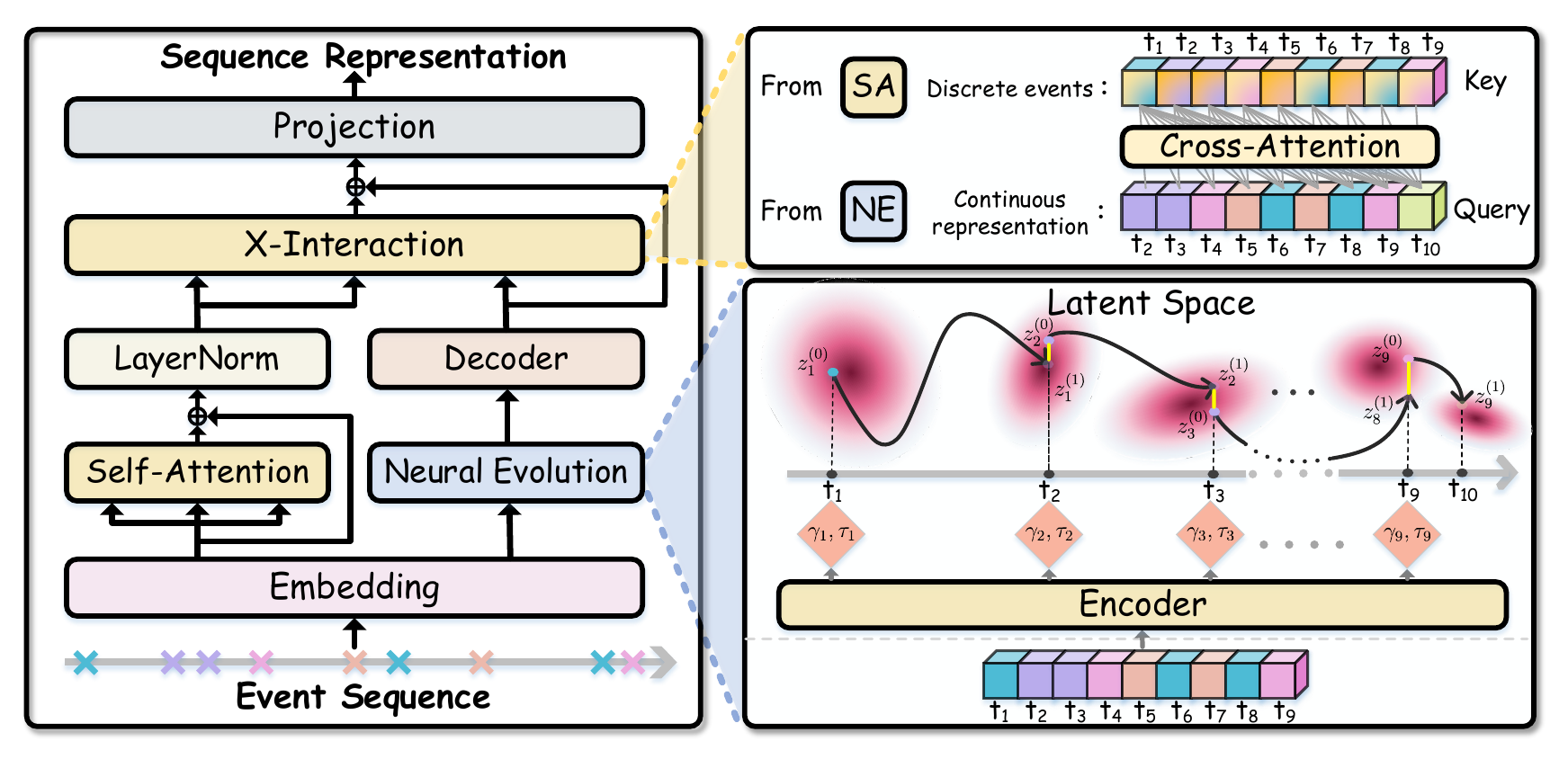} 
\caption{The overall framework NEXTPP. Left(SA): Discrete event sequence processing through embedding and self-attention layers. Right(NE): Continuous representation via latent space. X-Interaction(CA): Continuous-Discrete Interaction.}
\label{main_model}
\end{figure*}

\subsection{Sequence Representations}
Our NEXTPP pipeline proceeds in three stages: (1) a basic embedding layer mapping events into dense features, (2) a dual parallel Encoder via Self-Attention and Neural ODEs is employed to respectively extract the discrete-time temporal patterns and the continuous-time latent dynamics, (3) a Cross-Attention Fusion module integrates the two streams into a unified sequence representations. The overall architecture of our NEXTPP is illustrated in Figure \ref{main_model}.

\textbf{Embedding Layer.} 
The embedding layers are firstly employed to transform both event markers and timestamps into dense vectors. Specially, the event marker $m_i$ are projected into $\mathbf{Y}_{i} \in \mathbb{R}^D$ via an embedding matrix $\mathbf{M}\in {\mathbb R}^{|M|*D}$ \cite{xue2023easytpp}. For event times, we apply the position encoding scheme in \cite{zuo2020transformer}:
\begin{equation} \label{position_encoding}
[\mathbf{F}_i]_l = 
\begin{cases}
    \cos(t_i/10000^{\frac{l-1}{D}}), & \text{if } \text{l is odd,} \\
    \sin(t_i/10000^{\frac{l}{D}}), & \text{if } \text{l is even,}
\end{cases}
\end{equation}
as in Eq \eqref{position_encoding}, each timestamp \(t_i\) are encoded into the temporal embedding $\mathbf{F}_i \in {\mathbb R}^D$ via the trigonometric function with the embedding dimension \(D\), and $[\mathbf{F}_i]_l$ denotes the corresponding $l$-th component, where $l \in{1, 2, \cdots, D}$. The final representation of event $i$ is the element-wise sum $\mathbf{E}_i=\mathbf{Y}_i+\mathbf{F}_i$, so the entire sequence is thus compactly expressed as the embedding matrix $\mathbf{E}\in {\mathbb R}^{L*D}$.

\textbf{Neural Evolution.} We project each event  $\mathbf{E}_i$ into the low-dimensional manifold via latent space encoder:

\begin{equation} \label{encoder_mapping}
\begin{gathered}
    \bm{\gamma}_i = \mathcal{NN}_{enc\_\gamma} (\mathbf{E}_{i});\;\;
    \log\bm{\tau}_i = \mathcal{NN}_{enc\_\tau} (\mathbf{E}_{i}),
\end{gathered}
\end{equation}
where $\mathcal{NN}$ denotes a fully-connected neural network, $\bm{\gamma}_i \in \mathbb{R}^d$ and $log\bm{\tau}_i \in \mathbb{R}^d$ denote the mean and $log$-variance of latent distribution. The latent dimension $d$ satisfies $d \ll D$, enabling efficient simulation of the underlying dynamics.

Leveraging the reparameterization trick, the initial latent state for each event $e_i$ can be sampled as:
\begin{equation}\begin{gathered}
\mathbf{z}_{i}^{(0)}=\bm{\gamma}_i+\bm{\tau}_i\odot \bm{\epsilon},\quad\epsilon \sim \mathcal{N}(\textbf{0},\textbf{I})
\end{gathered}\end{equation}
then the latent state between a pair of events evolves over time according to the following Neural ODEs:
\begin{equation} \label{neural_ode}
    \frac{\mathrm{d} \mathbf{z}}{\mathrm{d} t}=f_{\theta}(\mathbf{z},t)
\end{equation}
where $f$ is parameterized by a linear network with learnable $\theta$, and the initial latent state follows the variational distribution $q(\mathbf{z}_i^{(0)}|t_i,m_i)$. For memory-efficient gradient computation, the continuous dynamic evolution is integrated by a black-box ODE solver with the adjoint sensitivity method.
\begin{equation}
    \mathbf{z}_{i}^{(1)}=\mathrm{ODESolve}(f_{\theta}, \mathbf{z}_{i}^{(0)},[t_{i},t_{i+1}])
\end{equation}

Finally, one-layer linear decoder maps the terminal latent state $\mathbf{z}_{i}^{(1)}$ to the the reconstructed embedding $\mathbf{O}_{i}$.
\begin{equation} \label{decoder_mapping}
\begin{gathered}
    \mathbf{O}_{i}=\mathcal{NN}_{dec}(\mathbf{z}_{i}^{(1)}).
\end{gathered}
\end{equation}

\textbf{X-Interaction.} To faithfully capture the influence of event history, we process the event embedding $\mathbf{E}_i$ in parallel with the neural evolution. First, a Self-Attention block distills an intermediate representation:
\begin{equation}
    \mathbf{A}_{i}=\mathrm{LayerNorm}({\mathbf{E}}_{i}+Self\text{-}Attention({\mathbf{E}}_{i})),
\end{equation}
which captures the intrinsic dependencies within the sequence. A Cross-Attention module then establishes bidirectional interaction between the discrete and continuous streams, where the reconstructed feature $\mathbf{O}_i$ from Neural ODE serves as the Query to attend over the intermediate representation $\mathbf{A}_i$ (as Key), enabling explicit information flow where historical event marks influence temporal dynamics while temporal context refines mark predictions:
\begin{equation}
    \mathbf{C}_{i}=\mathrm{LayerNorm}(\mathbf{O}_{i}+ \mathbf{X}\text{-}Interaction(\mathbf{O}_{i},\mathbf{A}_{i})),
\end{equation}
where $\mathbf{X}\text{-}Interaction$ employs Cross-Attention for implementation.
Eventually, The resulting $\mathbf{C}_{i}$ serves as the final sequence-level representation of our proposed NEXTPP.

\textbf{Intensity Function.} Armed with a high-fidelity sequence representation $\mathbf{C}$ of event sequence $H_{t_{i}}$, we express the conditional intensity of every event mark $m$ at time $t$ through the formulation of Hawkes Process:

\begin{equation} \label{hawkes_intensity}
\lambda^i(t,m) =\text{Act}(\alpha_m(t - t_i) + \mathbf{W}_m^\top \mathbf{C}_{i} + b_m + \beta_m),
\end{equation} 
where $\alpha_m$, $\mathbf{W}_m$, $b_m$, $\beta_m$ are learnable in intensity network. $\alpha_m$ is the triggering kernel governing the intensity decay rate for mark $m$ at time $t$. $\mathbf{W}_m$ and $b_m$ represent the weights and biases corresponding to the historical dependencies. $\beta_m$ describes the occurrence rate of current event in the absence of triggering events. The activation $\text{Act}(\cdot)$ is a $Scaled-Softplus$, defined as $ \text{Act}(\cdot) = \gamma_m \log(1 + \exp(\frac{\cdot}{\gamma_m}))$, with a learnable scale $\gamma_m>0$ to ensures strictly positive intensity values. This design endows the model to flexibly adjust the curvature of the intensity function, faithfully capturing the unique temporal dynamics of each mark $m$.

\subsection{Training}
For NEXTPP training, the loss is mainly consisted of the following three complementary objectives. Firstly, we minimize the negative log likelihood to learn the model parameters of intensity function in Eq \eqref{hawkes_intensity}, formulated as: 

\begin{equation} \label{ll}
\begin{aligned}
    \mathcal{L}_{MLE} &= -\ell(\mathcal{S})\\
    &=-\sum _{i=1}^{L}log\lambda^i(t_i,m_i)+\int_{t_1}^{t_L} \sum_{l=1}^{M}\lambda^i(s,l)ds
\end{aligned}
\end{equation}
Moreover, we infer the joint low-dimensional distribution underlying each event time $t_i$ and mark $m_i$ via variational inference. Concretely, we minimize the Kullback-Leibler(KL)\cite{blei2017variational} divergence between variational posterior and the true latent posterior for the entire observed sequence, written as:
\begin{equation}
\mathcal{L}_{KL}=\sum_{i=1}^{L}\mathcal{D}_{\mathrm{KL}}(q(\mathbf{z}_i^{(0)}\mid (t_i,m_i))\mid\mid p(\mathbf{z}_i^{(0)})). 
\end{equation}

where $q(\mathbf{z}_i^{(0)}| (t_i,m_i)) \sim \mathcal{N}(\bm{\gamma}_i, \bm{\tau}_i)$ and $p(\mathbf{z}_{i}^{(0)}) \sim \mathcal{N}(\textbf{0},\textbf{I})$. 
Since event times arise from a continuous domain, we add a third loss term encouraging smooth latent space trajectories\cite{higgins2017beta}. we penalize the discrepancy between the evolved representation of the current event $\mathbf{z}_{i}^{(1)}$ driven by $\mathbf{z}_{(i)}^{(0)}$ and the static representation of the next observed event $\mathbf{z}_{(i+1)}^{(0)}$, thereby preserving latent space trajectory continuity across the sequence:

\begin{equation}
\mathcal{L}_{cont}=\sum_{i=1}^{L-1}(\mathbf{z}_{i}^{(1)}-\mathbf{z}_{(i+1)}^{(0)})^2.
\end{equation}

Finally, our overall loss function can be expressed as:
\begin{equation}
\mathcal{L}=\mathcal{L}_{MLE}+\mathcal{L}_{KL}+\mathcal{L}_{cont}.
\end{equation}

\subsection{Thinning Sampler for NEXTPP}

NEXTPP employs iterative thinning sampling\cite{lewis1979simulation} to simulate marked event sequences adhering to the conditional intensity in Eq \eqref{hawkes_intensity}. Each candidate event is accepted with probability preserving both temporal and mark dynamics:
\begin{enumerate}
    \item Initial a proposal process with the upper bound intensity over all possible marks at each event timestamp $t$: $ \lambda_{m}^{*}=\max\limits_{t \in (t_{i-1}, t_{i-1} + \Delta T_{\text{max}})} \lambda^{i}(t, m)+\epsilon$;

    \item Generate candidate time interval with the intensity in Step 1 from the homogeneous Poisson process: $ t_{m}^*= {\lambda}_m^{*}  exp(-{\lambda}_m^{*}(t-t_{i-1})) $;

    \item Accept $t_{i-1}+t_{m}^{*}$ as the sampling time $t_{i, m}$ with condition $\mu {\lambda}_m^{*} \leq \lambda^i(t_m^*, m)$ if it satisfies $t_{m}^{*}<\Delta T_{\text{max}}$, otherwise $t_{i, m}=\infty$;

\end{enumerate}
repeat Step 2-3 to obtain event time $t_i$ and mark $m_i$:
\begin{equation}
    t_i = \min\limits_{m} t_{i,  m},\; m_i = \arg\min\limits_{m} t_{i, m}.
\end{equation}
Through $L$ samplings iterations, we obtained a new sequence of events.

\section{Experiments}

\begin{table*}[t]
\centering
\setlength{\tabcolsep}{6pt} 
\caption{Dataset statistics with number of marks (M), event tokens distribution across splits, and sequence length characteristics.}
\begin{tabular}{c|c|ccc|ccc}
\toprule
\multirow{2}{*}{\textbf{Dataset}} & 
\multirow{2}{*}{\textbf{M}} & 
\multicolumn{3}{c|}{\textbf{Event Tokens}} & 
\multicolumn{3}{c}{\textbf{Sequence Length}} \\
 \cmidrule(lr){3-5} \cmidrule(lr){6-8}
 & 
 & \textbf{Train} & \textbf{Dev} & \textbf{Test} & \textbf{Min} & \textbf{Mean} & \textbf{Max} \\
\midrule
RETWEET       & 3  & 369731 & 62823 & 61154 & 10  & 41  & 97 \\
AMAZON        & 16 & 288377 & 40995 & 84048 & 14  & 44  & 94 \\
TAXI          & 10 & 51854 & 7404 & 14820 & 36  & 37  & 38 \\
STACKOVERFLOW & 22 & 90497 & 25762 & 26518 & 41  & 65  & 101 \\
EARTHQUAKE    & 7  & 49363 & 6612 & 14748 & 15  & 16  & 18 \\
\bottomrule
\end{tabular}
\label{dataset}
\end{table*}

\subsection{Experimental Settings}

\paragraph{Datasets} We evaluate the model performance on five real-world datasets (\textbf{Taxi}, \textbf{Amazon}, \textbf{StackOverflow}, \textbf{Earthquake}, \textbf{Retweet}) and follow the pre-processing procedure from \cite{xue2023easytpp}. The core information of each dataset is as follows: The Taxi \cite{whong2014foiling} dataset is derived from taxi pickup and drop-off records in New York City, containing sequence data of 2,000 drivers with a total of 10 marks (representing combinations of different boroughs and pickup/drop-off statuses); The Amazon  \cite{NiDataset2018} dataset covers user product review records from January 2008 to October 2018, selecting sequences of 5,200 active users (with an average of 70 events per user) and using product categories as 16 marks; The StackOverflow \cite{jure2014snap} dataset is from two years of badge award records on this Q\&A platform, including sequences of 2,200 users and taking 22 badge types as marks; The Earthquake \cite{xue2023easytpp} dataset records earthquake events with magnitudes $ \geq $ 2.5 in the United States from 1996 to 2023, which are divided into 7 intervals by magnitude as marks, and the data is sourced from the U.S. Geological Survey \footnote{https://earthquake.usgs.gov/earthquakes/search/}; The Retweet \cite{zhou2013learning} dataset contains 5,200 tweet sequences (each sequence includes an original tweet and subsequent retweets), with the follower count levels of retweeters (small/medium/large) serving as 3 marks.Table \ref{dataset} summarizes the statistical information of the processed datasets.

\paragraph{Baselines} We evaluate our model by comparing it with state-of-the-art baselines: \textbf{(i) Discrete-time approximation models}: classical \textbf{Multivariate Hawkes Process (MHP)}, \textbf{Recurrent Marked Temporal Point Process (RMTPP)} \cite{du2016recurrent}, \textbf{Neural Hawkes Process (NHP)} \cite{mei2017neural}, \textbf{Transformer Hawkes Process (THP)} \cite{zuo2020transformer},  and \textbf{Self-Attentive Hawkes Process (SAHP)} \cite{zhang2020self}, \textbf{Attentive Neural Hawkes Process (AttNHP)} \cite{yang2022transformer}; \textbf{(ii) Continuous-time modeling models}: \textbf{IFTPP} \cite{rubanova2019latent}, \textbf{ODETPP} \cite{xue2023easytpp}, and \textbf{DLTPP} \cite{zhou2025non}.

\begin{table}[t]
  \centering
  \fontsize{10pt}{8.5pt}\selectfont
  \caption{Results of all model reproductions across five real datasets. The performance of our NEXTPP model represents the average of five training runs with different random seeds. Lower scores indicate better performance. The best results are indicated in \textbf{bold} and suboptimal results are \underline{underlined}.(R = RMSE ($\downarrow$), E = Error Rate \% ($\downarrow$), S.O. = StackOverflow, EQ = Earthquake)}
  \renewcommand{\arraystretch}{1.5}
  \setlength{\tabcolsep}{3pt}
  \begin{tabular}{@{}lcccccccccc@{}}
    \toprule
    \multirow{2}{*}{\textbf{Method}} & 
    \multicolumn{2}{c}{\textbf{Taxi}} & 
    \multicolumn{2}{c}{\textbf{Amazon}} & 
    \multicolumn{2}{c}{\textbf{S.O.}} & 
    \multicolumn{2}{c}{\textbf{EQ}} & 
    \multicolumn{2}{c}{\textbf{Retweet}} \\
    \cmidrule(lr){2-3} \cmidrule(lr){4-5} \cmidrule(lr){6-7} \cmidrule(lr){8-9} \cmidrule(l){10-11}
    & R & E & R & E & R & E & R & E & R & E \\
    \midrule
    MHP     & 0.382  & 9.53  & 0.635  & 75.9  & 1.388  & 65.0  & 4.128  & 54.9  & 22.920 & 55.7  \\
    RMTPP   & 0.374  & 10.4  & 0.474  & 68.1  & 1.392  & 56.3  & 2.097  & 54.9  & 26.508 & 45.8  \\
    NHP     & 0.373  & 10.0  & 0.513  & 68.5  & 1.431  & 53.3  & 2.347  & 54.7  & 22.410 & 39.9  \\
    THP     & 0.370  & 10.1  & 0.471  & 66.6  & 1.343  & 53.3  & 2.160  & 55.1  & 24.491 & 39.9  \\
    SAHP    & \underline{0.334} & 10.0  & 0.549  & 68.9  & \underline{1.331} & 56.5  & 1.864  & 54.9  & 21.673 & 40.6  \\
    AttNHP  & 0.394  & 14.1  & 0.652  & 67.3  & 1.402  & 53.7  & 2.117  & 54.8  & 21.748 & 40.9  \\
    \midrule
    IFTPP   & 0.404  & \underline{9.38} & \underline{0.461} & \textbf{65.6} & 1.572  & \underline{52.9} & 2.540  & \textbf{53.1} & 34.302 & \underline{39.8} \\
    ODETPP  & 0.359  & 11.6  & 0.671  & 68.5  & 1.507  & 54.2  & 2.396  & 56.0  & \underline{19.404} & 57.6  \\
    DLTPP   & 0.662  & 14.9  & 0.485  & 66.9  & 1.497  & 59.5  & \underline{1.672} & 63.4  & 21.892 & 42.9  \\
    \midrule
    NEXTPP  & \textbf{0.323} & \textbf{8.83} & \textbf{0.377} & \underline{66.5} & \textbf{1.152} & \textbf{52.2} & \textbf{1.542} & \underline{54.4} & \textbf{19.340} & \textbf{35.5} \\
    \bottomrule
  \end{tabular}
  \label{performance}
\end{table}

\begin{table}[t]
\centering
\fontsize{9pt}{10.5pt} \selectfont
\renewcommand{\arraystretch}{1.5} 
\setlength{\tabcolsep}{2pt} 
  \caption{Log-Likelihood Performance Comparison. Reported values are the mean log-likelihood over five independent runs, with the corresponding standard deviation indicating experimental variability.}
  \begin{tabular}{@{}lccccc@{}} 
    \toprule
    \multirow{3}{*}{MODEL} & 
    \multicolumn{5}{c}{Log-Likelihood} \\
    \cmidrule(l){2-6} 
    & Taxi & Amazon & \begin{tabular}{@{}c@{}}Stack\\Overflow\end{tabular} & Earthquake & Retweet \\
    \midrule
RMTPP &  $\text{0.219}_{\pm \text{0.033}}$ & $\text{-2.258}_{\pm \text{0.009}}$  & $\text{-2.820}_{\pm \text{0.025}}$  & $\text{-2.313}_{\pm \text{0.272}}$  & $\text{-5.754}_{\pm \text{1.785}}$ \\
NHP &  $\text{0.440}_{\pm \text{0.007}}$ & $\text{-2.036}_{\pm \text{0.066}}$  & $\text{-2.440}_{\pm \text{0.103}}$  & $\text{-2.155}_{\pm \text{0.056}}$  & $\text{-4.020}_{\pm \text{0.133}}$ \\
THP &  $\text{0.217}_{\pm \text{0.034}}$ & $\text{-2.215}_{\pm \text{0.016}}$  & $\text{-2.478}_{\pm \text{0.031}}$  & $\text{-2.353}_{\pm \text{0.040}}$  & $\text{-4.143}_{\pm \text{0.005}}$ \\
SAHP & $\text{-0.295}_{\pm \text{0.266}}$ & $\text{-2.417}_{\pm \text{0.207}}$  & $\text{-6.930}_{\pm \text{0.007}}$  & $\text{-3.136}_{\pm \text{0.610}}$  & $\text{-4.798}_{\pm \text{0.637}}$ \\
AttNHP &  $\text{0.164}_{\pm \text{0.031}}$ & $\text{-1.364}_{\pm \text{0.195}}$  & $\text{-2.560}_{\pm \text{0.004}}$  & $\text{-2.359}_{\pm \text{0.026}}$  & $\text{-4.128}_{\pm \text{0.002}}$ \\\midrule
NEXTPP &  $\textbf{0.586}_{\pm \textbf{0.026}}$ & $\textbf{-1.246}_{\pm \textbf{0.071}}$  & $\textbf{-1.971}_{\pm \textbf{0.032}}$  & $\textbf{-1.828}_{\pm \textbf{0.034}}$  & $\textbf{-3.993}_{\pm \textbf{0.007}}$ \\
    \bottomrule
  \end{tabular}
  \label{ll_performance}
\end{table}

\begin{table}[t]
\centering
\fontsize{8pt}{7pt} \selectfont 
\renewcommand{\arraystretch}{1.2} 
\setlength{\tabcolsep}{0.5pt}
  \caption{Ablation results on Amazon, StackOverflow and Retweet.}
  \begin{tabular}{@{}c|ccc|ccc|ccc} 
    \toprule
    \multirow{2}{*}{Settings} & 
      \multicolumn{3}{c}{Amazon} & 
      \multicolumn{3}{c}{StackOverflow} &
      \multicolumn{3}{c}{Retweet}\\
    \cmidrule(lr){2-4} \cmidrule(lr){5-7}  \cmidrule(lr){8-10}  
       & LL($ \uparrow $)& RMSE($ \downarrow $)& ErrRt($ \downarrow $)&  LL($ \uparrow $)& RMSE($ \downarrow $)& ErrRt($ \downarrow $) & LL($ \uparrow $)& RMSE($ \downarrow $)& ErrRt($ \downarrow $)\\
    \midrule
    w/o NE                  & -2.178 & 0.468 & 66.8\% & -2.474 & 1.338 & 53.3\% 
                            &-4.180 & 20.541 & 40.2\%\\ 
    w/o CA                   & -1.530 & 0.438 & 71.1\% & -2.657 & 2.199 & 56.4\%
                            &-4.122 & 20.108 & 47.3\%\\ 
    \midrule

    NE$\rightarrow$GRU      & -2.235 & 0.481 & 67.7\% & -2.369 & 1.366 & 53.2\% 
                            &-4.166 & 21.145 &40.7\%\\ 
    NE$\rightarrow$LSTM      & -2.263 & 0.495 & 68.2\% & -2.403 & 1.389 & 53.3\%
                            &-4.177 & 20.795 &40.6\%\\ 
    \midrule
    NEXTPP                  & \textbf{-1.246} &\textbf{ 0.377} & \textbf{66.5\%} & \textbf{-1.971} & \textbf{1.152} & \textbf{52.2\%} & \textbf{-3.993} & \textbf{19.340} & \textbf{35.5\%}\\ 
    \bottomrule
  \end{tabular}

  \label{Ablation}
\end{table}

\paragraph{Evaluation Metrics}To comprehensively evaluate NEXTPP, we use three metrics to evaluate the model's performance from three aspects: Compute the log-likelihood to evaluate the model's \textbf{goodness-of-fit} to the overall distribution: $\ell(S)=\sum _{i=1}^{L}\log\lambda^i(t_i,k_i)-\int_{0}^{T}\sum_{l=1}^{M}\lambda^i(\tau,l)d\tau$. We quantify \textbf{temporal prediction} accuracy via: $RMSE = \sqrt{\frac{1}{n}\sum_{i=1}^{n}(t_i - \hat{t}_i)^2}$. To evaluate the model's \textbf{type prediction} accuracy, we define the error rate as: $Error Rate = \frac{\text{FP} + \text{FN}}{\text{TP} + \text{TN} + \text{FP} + \text{FN}}$.

\paragraph{Implementation Details} Our method is implemented in Pytorch on a single NVIDIA GeForce RTX 4090 GPU with 24 GB memory. We set the learning rate as 1e-3 via the searching in a set of \{1e-4, 5e-4, 1e-3\}. We utilized the Adam optimizer and set consistent hyperparameters across most datasets: hidden size = 128, time embedding dimension = 16, and dropout rate = 0.1. For the number of layers  and training epochs , we tailored them to each dataset: Taxi, StackOverflow, and Retweet used 2 layers with 150 epochs; Amazon used 1 layer with 50 epochs; and Earthquake used 4 layers with 150 epochs.

\subsection{Overall Performance}

Our primary results are presented in Table \ref{performance}, which reports event-type error and temporal RMSE, and Table \ref{ll_performance}, which presents the log-likelihood of observed event sequences under the learned model.
\begin{itemize}
    \item \textbf{Overall Performance Validation. } Across five diverse datasets, NEXTPP achieves the lowest RMSE on every benchmark, reducing next-timestamp error from 0.461 to 0.377 on Amazon and from 1.331 to 1.152 on Stack Overflow. While NEXTPP also attains the smallest event-type error in three out of five cases (Taxi, Stack Overflow, Retweet). On the two datasets where classification error is marginally higher, we observe that IFTPP exhibits severe performance fluctuations on specific datasets with respect to temporal RMSE. In contrast, NEXTPP maintains consistent performance, demonstrating a robust balance between temporal precision and mark accuracy.
    \item \textbf{Superiority of Continuous-Time Distribution Learning. } By leveraging Neural ODEs for continuous evolution, NEXTPP consistently surpasses all baselines in log-likelihood, demonstrating a markedly better fit to the empirical event distributions. These gains confirm that modeling the latent trajectory as a smooth, event-driven ODE yields a more accurate density estimation than discrete or intensity-parametric alternatives.
\end{itemize}

\subsection{Ablation Studies}
We conduct comprehensive ablations to quantify the contribution of each architectural component in NEXTPP. 
Specifically, we evaluate: (i) the effect of omitting the Neural Evolution module (w/o NE) and the Cross-Attention module (w/o CA); and (ii) the impact of replacing our Neural ODE backbone in the Neural Evolution module with two recurrent alternatives, GRU and LSTM. Results are summarized in Table \ref{Ablation}, leading to two key findings: (1) \textbf{Module-Level Contributions.} Every component materially improves performance. Excluding Cross-Attention increases the event-type error rate by more than 4\% points on average, underscoring its role in integrating context across event streams. Likewise, removing Neural Evolution causes a pronounced degradation in log-likelihood, confirming its importance for learning a faithful distributional representation. (2) \textbf{Advantages of Continuous-Time Dynamics.} Substituting the ODE solver with GRU or LSTM yields modest rises in error rate but a larger drop in log-likelihood. This contrast reflects the Neural ODE’s ability to model continuous-time trajectories with fine-grained temporal precision, which recurrent architectures approximate only at discrete steps.

\begin{figure*}[t]
\centering
\includegraphics[width=0.83\textwidth]{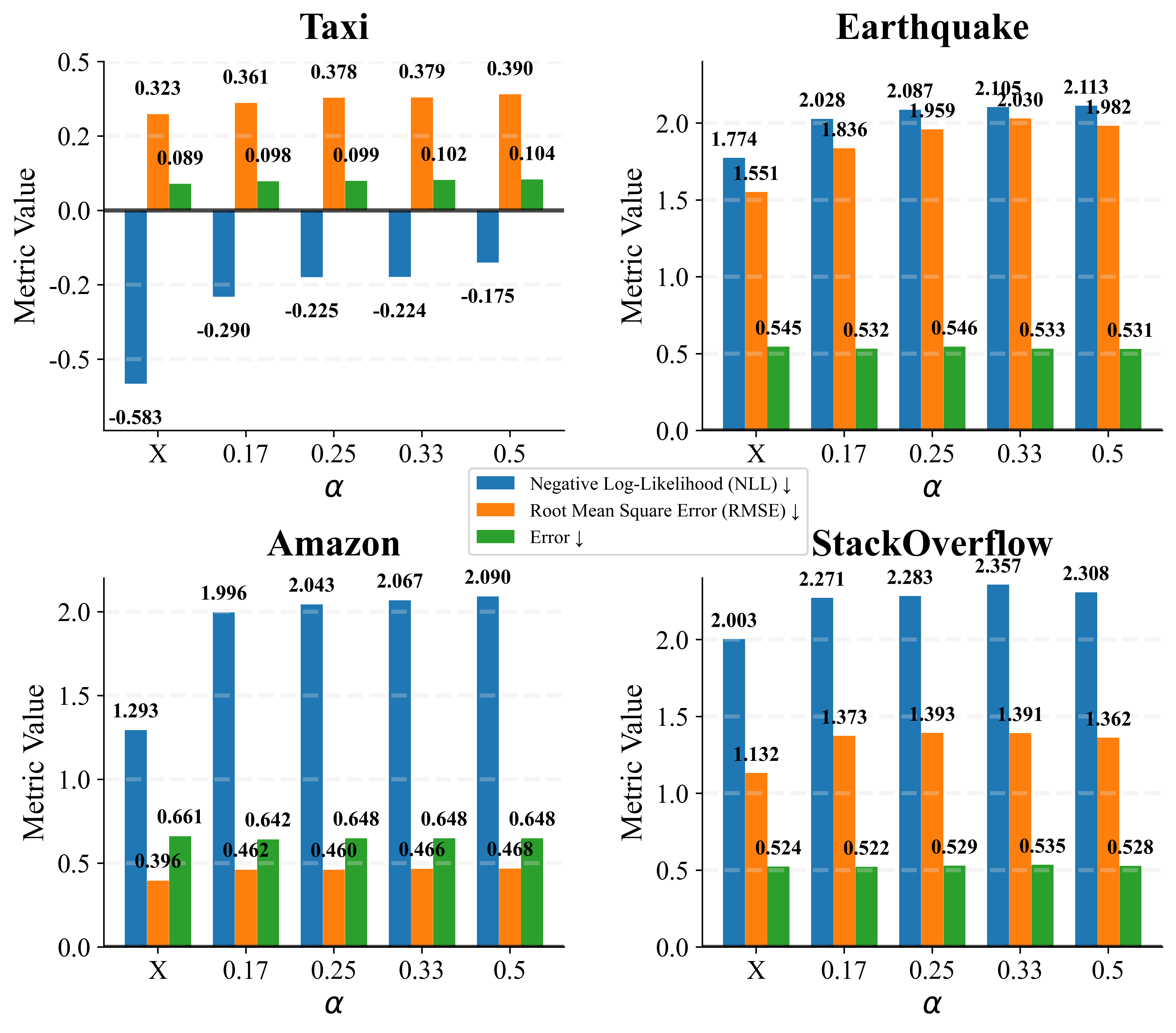} 
\caption{Block-wise test results: The sequence is divided into several blocks according to partition ratio $ \alpha $, with each block evolving in its corresponding latent space, where X indicates one block per event.(Lower scores are better)}
\label{chunk_bar_TE}
\end{figure*}

\subsection{Event-Granularity Evolution Validation}
To probe NEXTPP’s event-granularity evolution mechanism, we vary the event-to-event modeling granularity by partitioning each sequence into fixed-length chunks of 1/$\alpha$ events. As illustrated in Figure \ref{chunk_bar_TE}, among several datasets tested, the evolution divided by a single event performs better. The negative log-likelihood is almost monotonically increasing, indicating that coarse updates progressively undermine the model's capacity to approximate the true event distribution. Simultaneously, the RMSE curve trends upward, demonstrating lost precision in predicted timestamps when multiple inter-event transitions are merged into one latent step. This pattern indicates that larger chunks group multiple transitions into a single latent update can obscure the temporal dynamic characteristics at the micro level. Under this coarse regime, the model loses sensitivity to rapid changes in event intensity and cannot adapt to varying inter-event intervals, which in turn degrades both its fit to the data and its forecasting precision. In contrast, single-event granularity compels the Neural ODE solver to register each inter-event derivative, faithfully reflecting abrupt spikes, gradual decays, and oscillatory patterns intrinsic to the dynamics. By maintaining every event transition as an individual state, NEXTPP captures subtle timing cues that inform both density estimation and label prediction.

\begin{figure}[t]
	\centering
	\raisebox{6\height}[0pt][0pt]{
		\makebox[0pt][r]{\textbf{(a)}\hspace*{-0.3 em}}
	}%
	\subfloat{
		\includegraphics[scale=0.1]{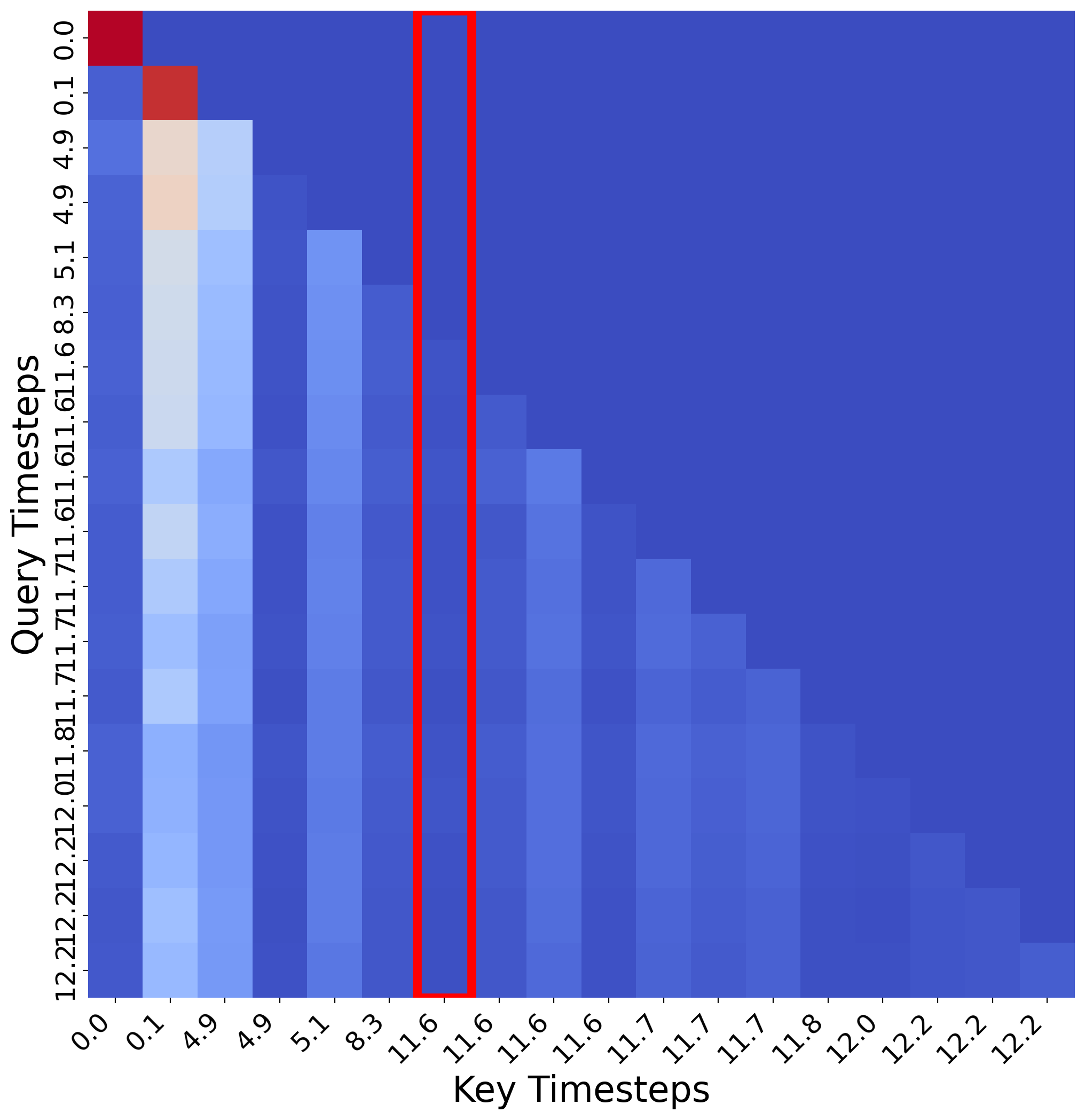}
	}%
         \subfloat{		
		\includegraphics[scale=0.1]{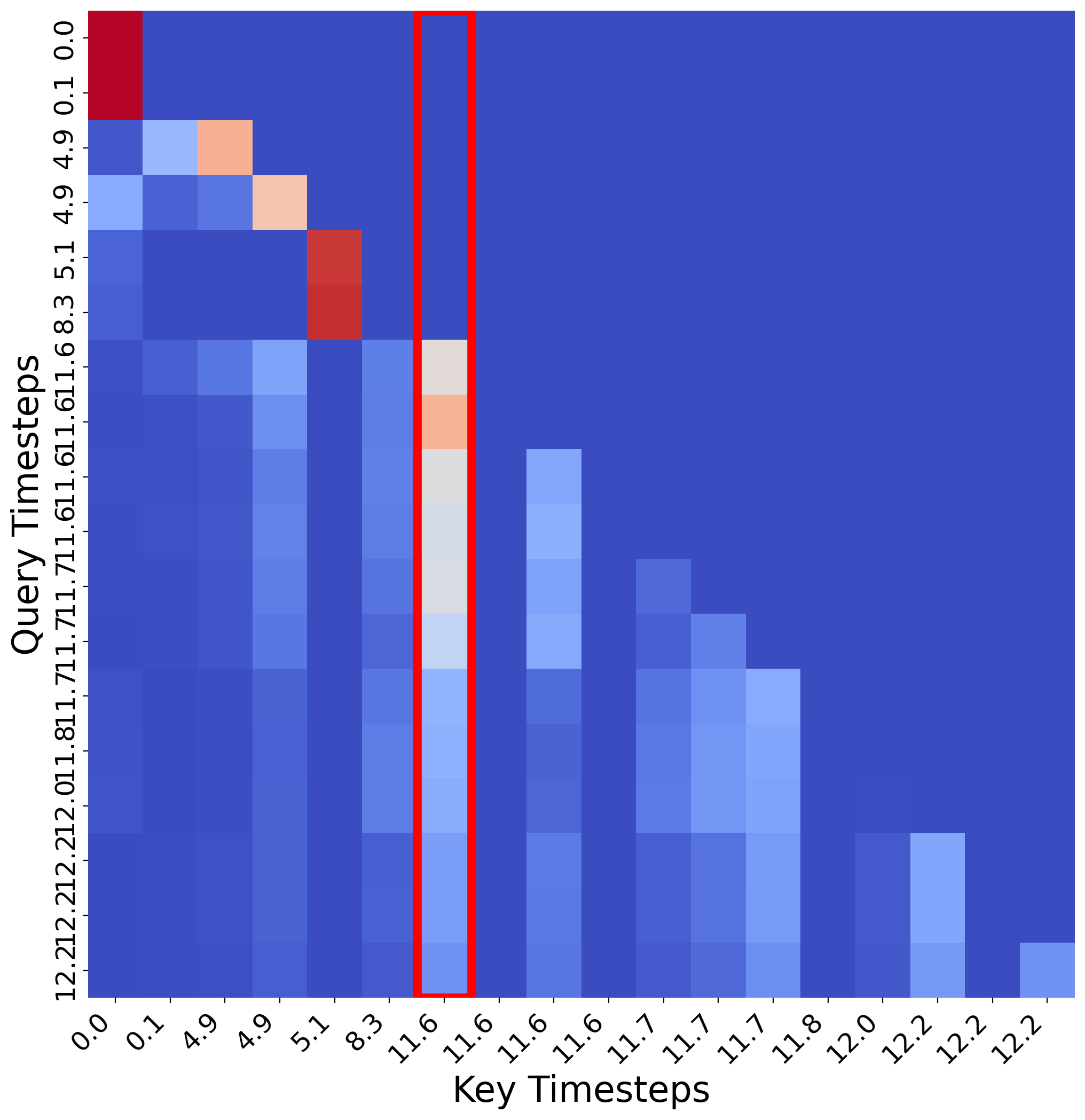}
	}%
          \subfloat{		
		\includegraphics[scale=0.1]{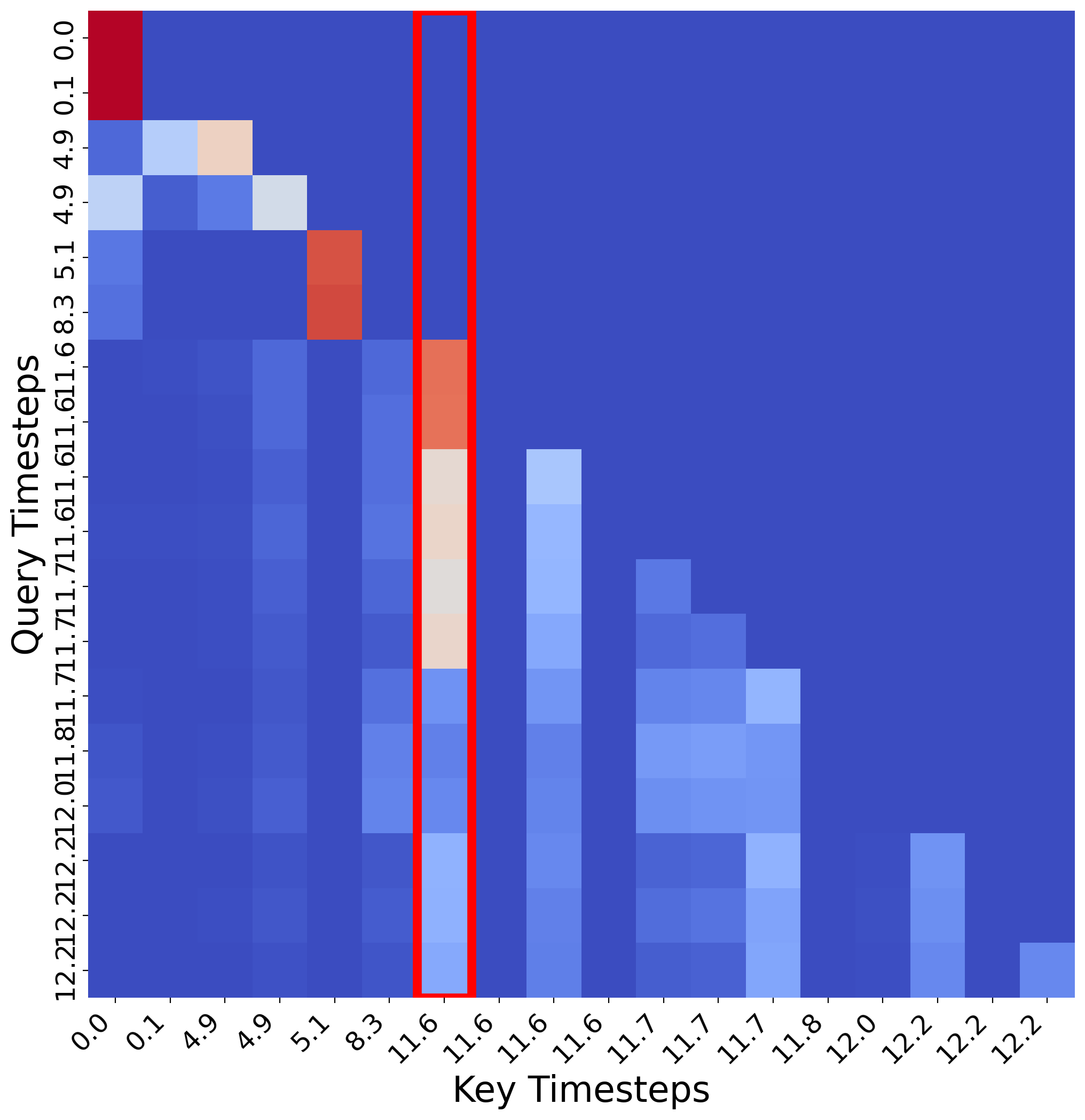}
	}%
          \subfloat{		
		\includegraphics[scale=0.1]{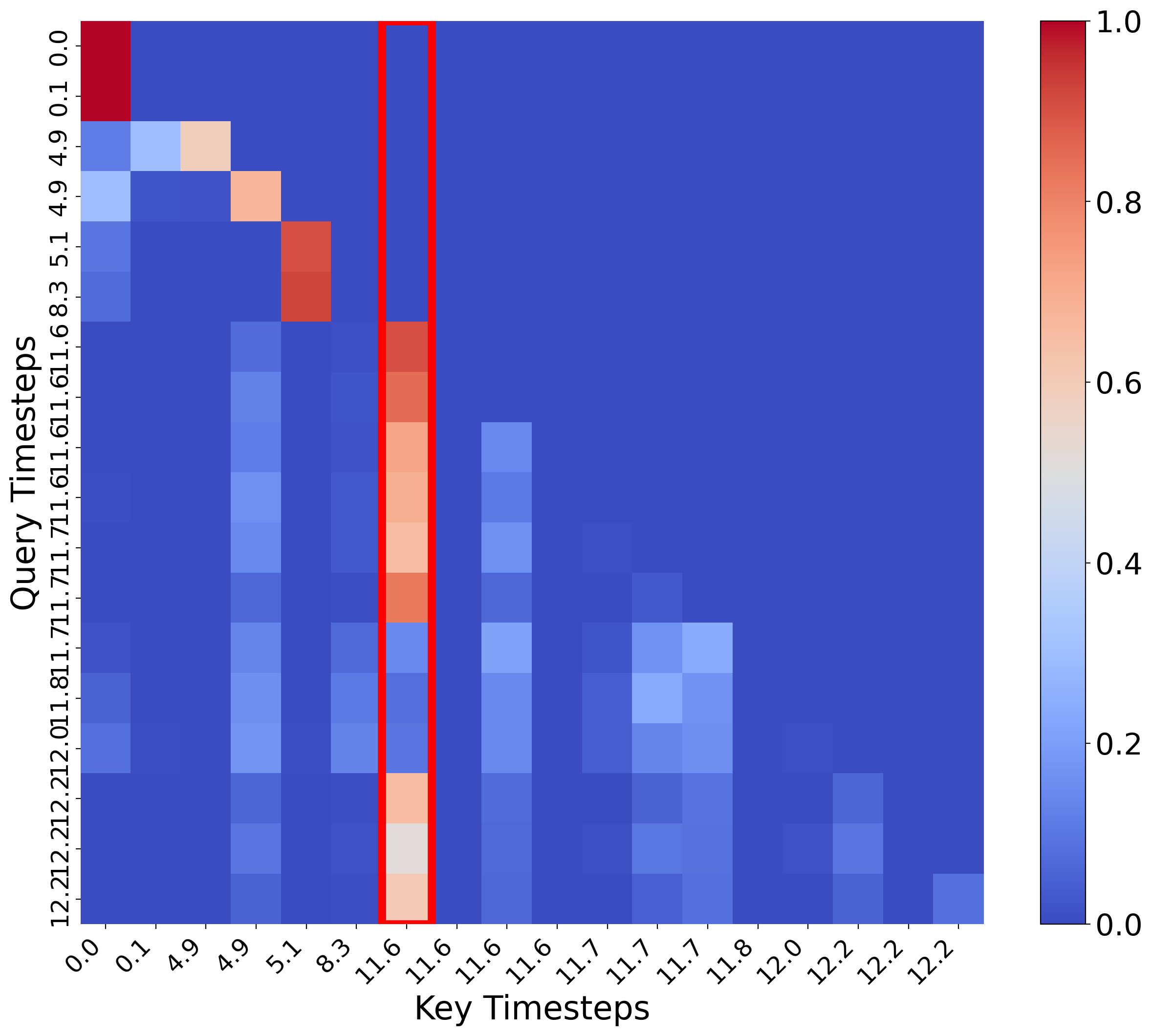}
	}%
 \\
	\raisebox{6\height}[0pt][0pt]{
		\makebox[0pt][r]{\textbf{(b)}\hspace*{-0.3 em}}
	}%
	\subfloat{
		\includegraphics[scale=0.1]{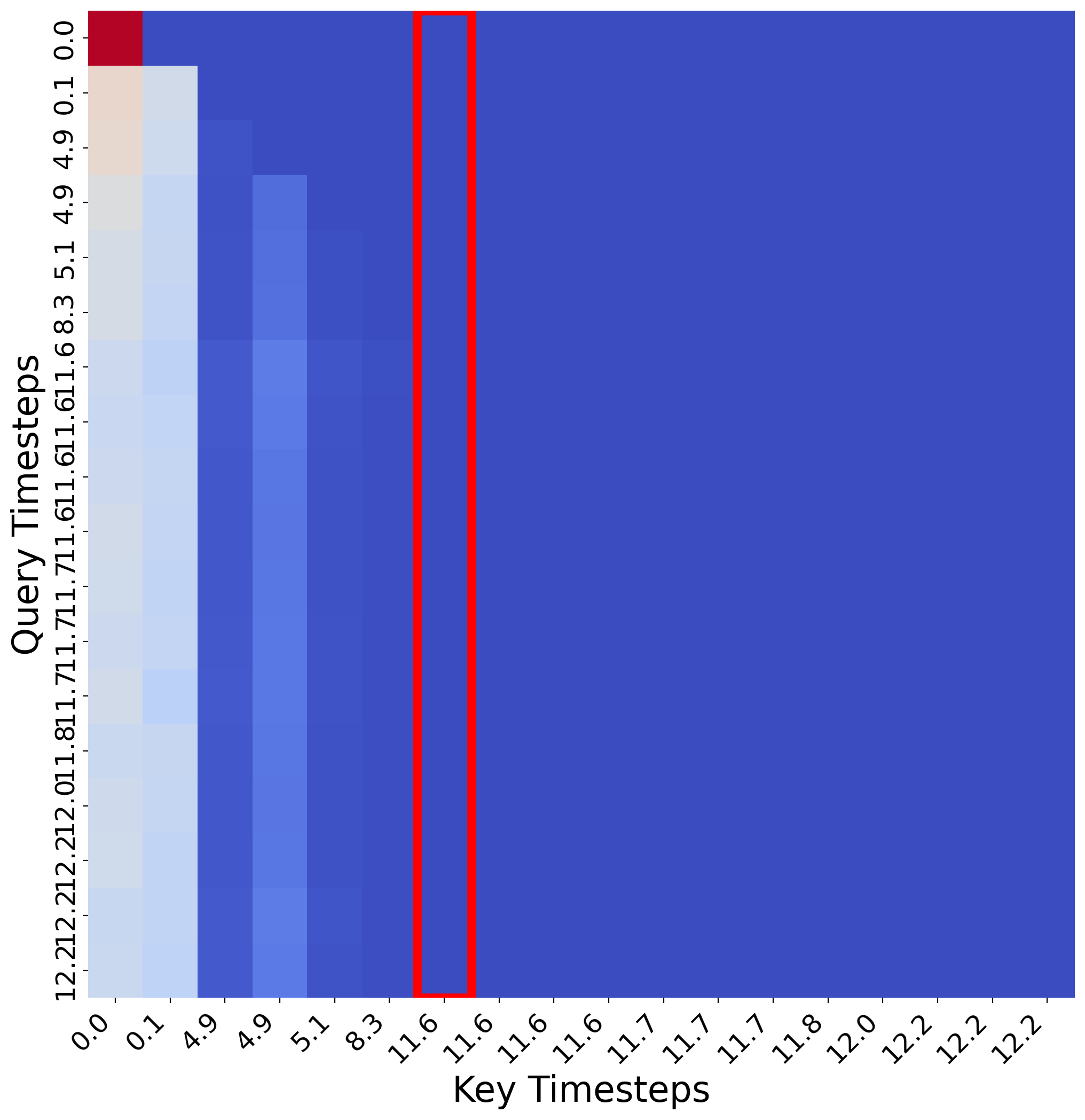}
	}%
         \subfloat{		
		\includegraphics[scale=0.1]{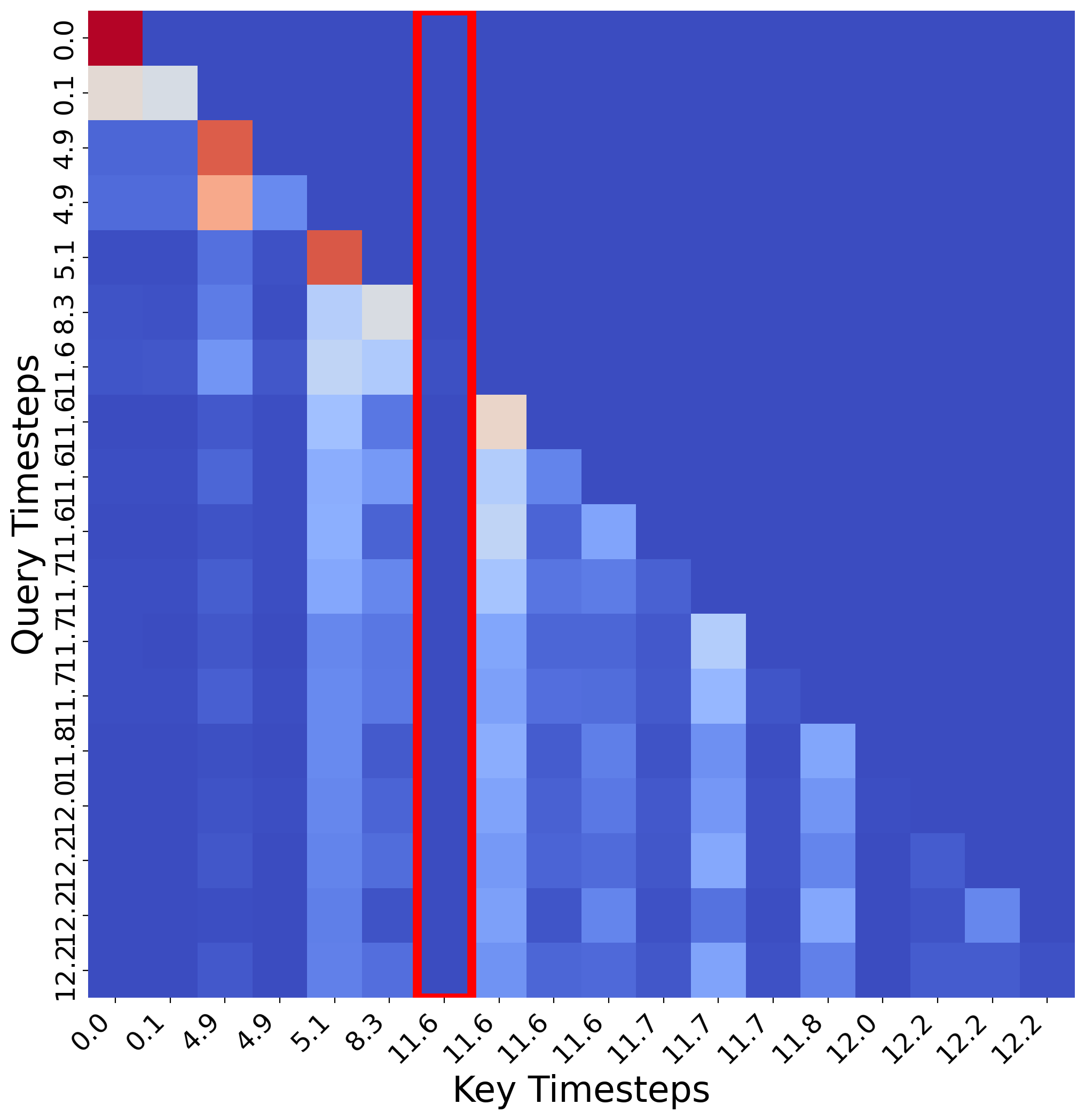}
	}%
          \subfloat{		
		\includegraphics[scale=0.1]{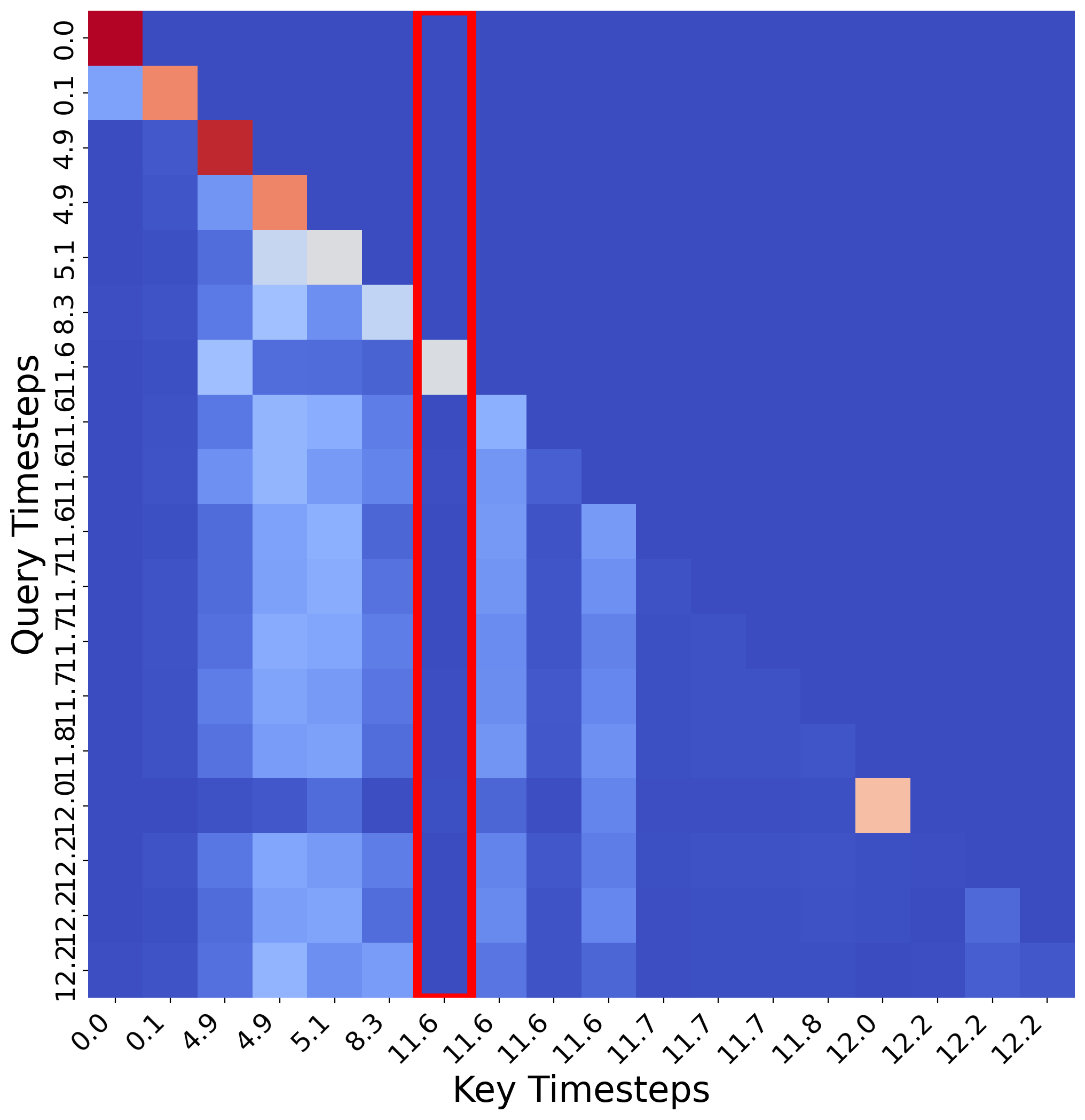}
	}%
          \subfloat{		
		\includegraphics[scale=0.1]{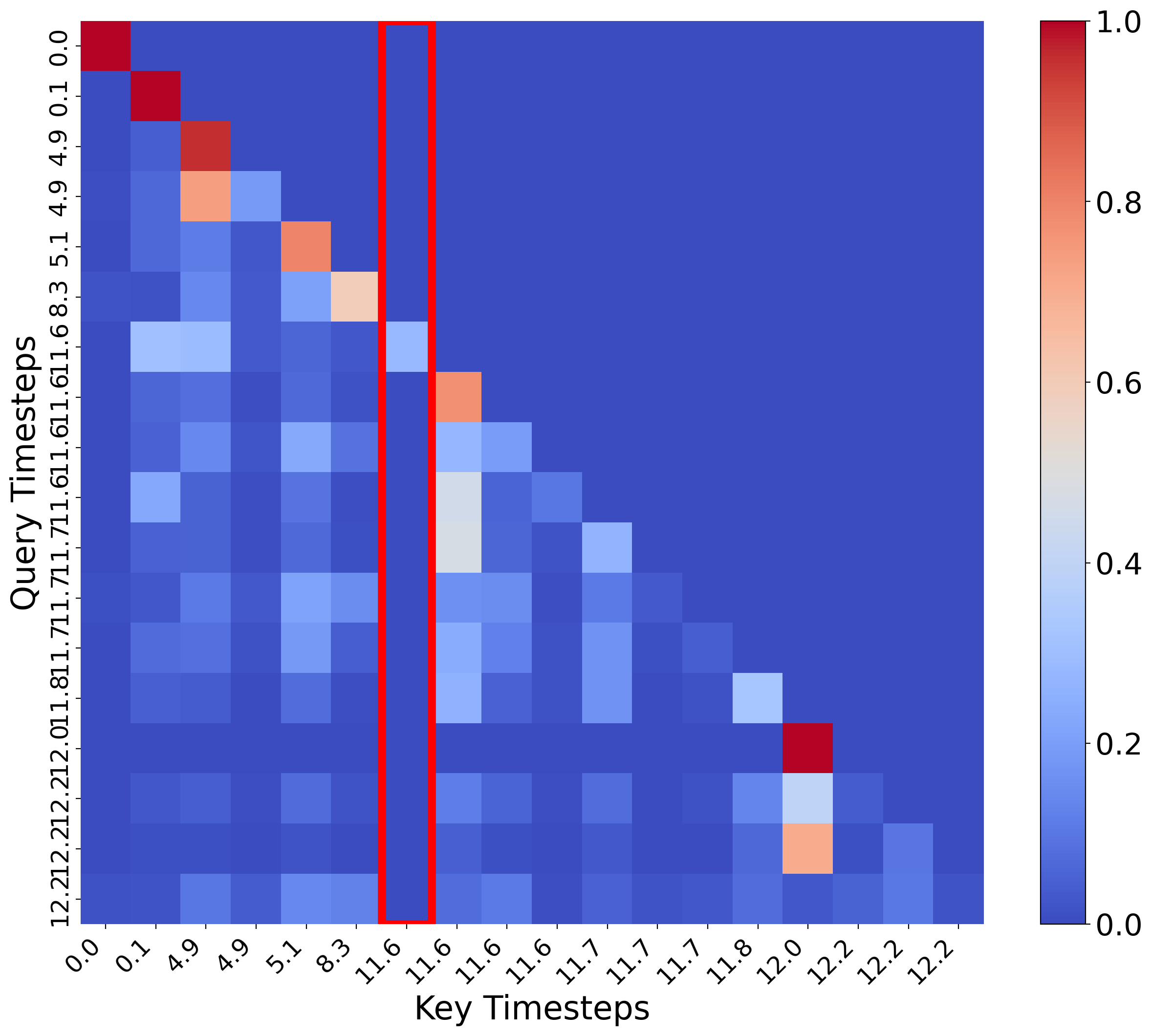}
	}%
    \medskip \\
    \begin{tabular}{@{}*{5}{c}@{}}
        Epoch: 0 \;  \; \; \; \; \; \; &Epoch: 40 \; \; \; \; \; \; \; & Epoch: 70  \; \; \; \; \; \; \; & Epoch: 130 \\
    \end{tabular}
	\caption{Heatmap comparison of the seismic sequences (shown in Figure \ref{intro}) during model  training. (a) The first row shows our Continuous-Discrete Interaction($\mathbf{X}\text{-}Interaction$); (b) The second row displays the conventional Attention($Self\text{-}Attention$).}
	\label{ca1}
\end{figure}

\begin{figure}[ht]
\centering
\subfloat[LL(EQ)]{%
    \includegraphics[width=0.233\linewidth]{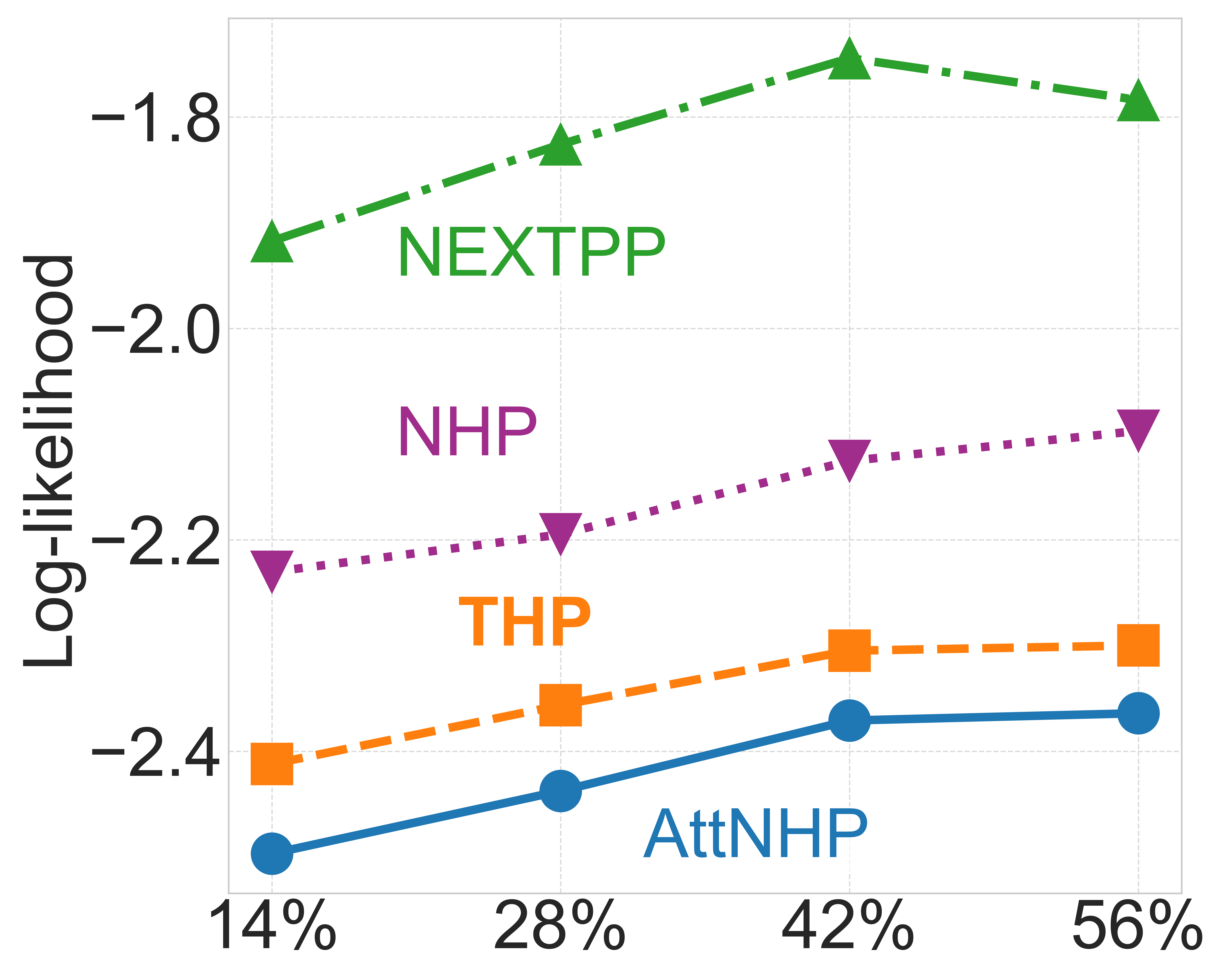}
}
\hfill
\subfloat[\small ErrRt(EQ)]{%
    \includegraphics[width=0.233\linewidth]{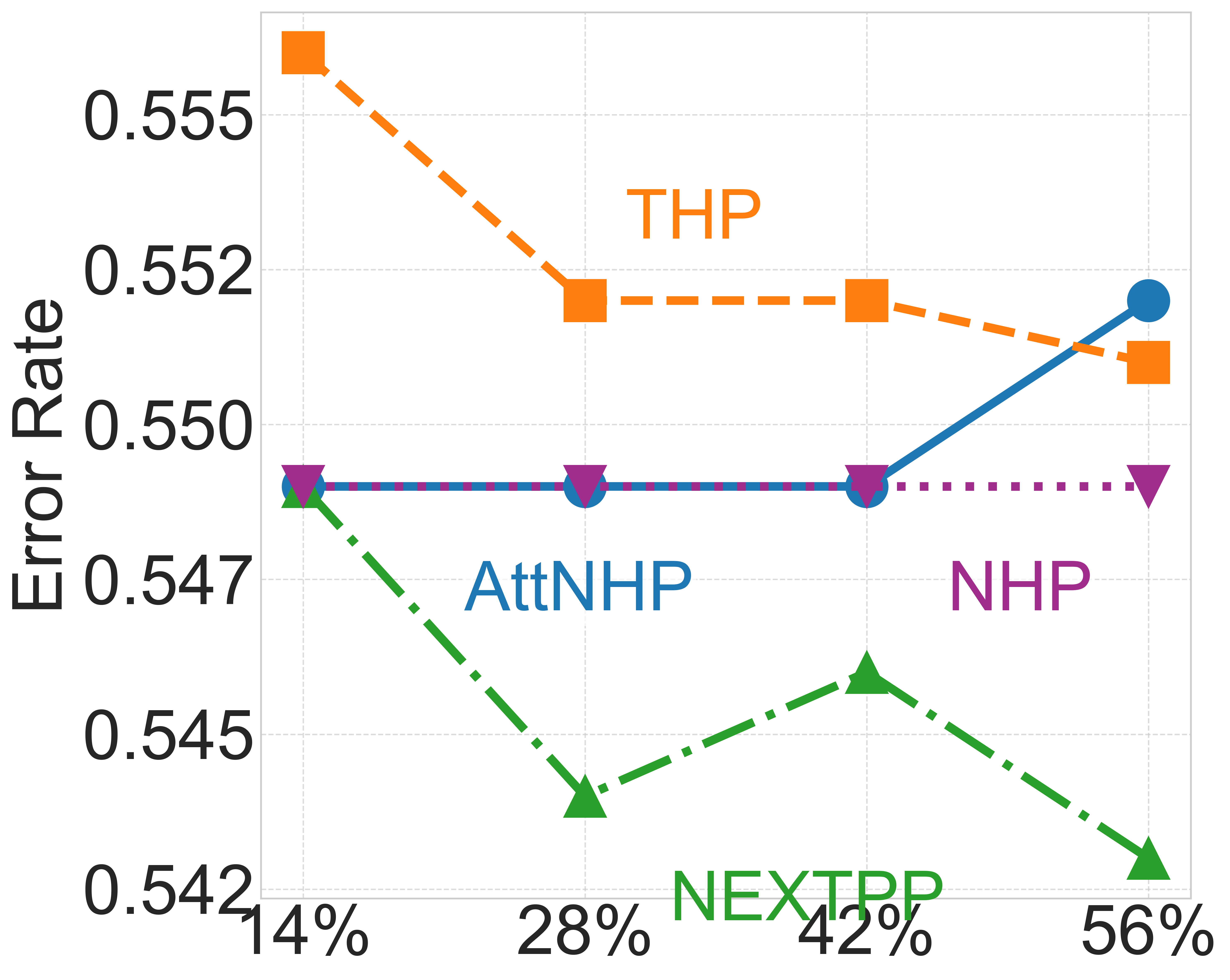}
}
\hfill
\subfloat[LL(Retweet)]{%
    \includegraphics[width=0.233\linewidth]{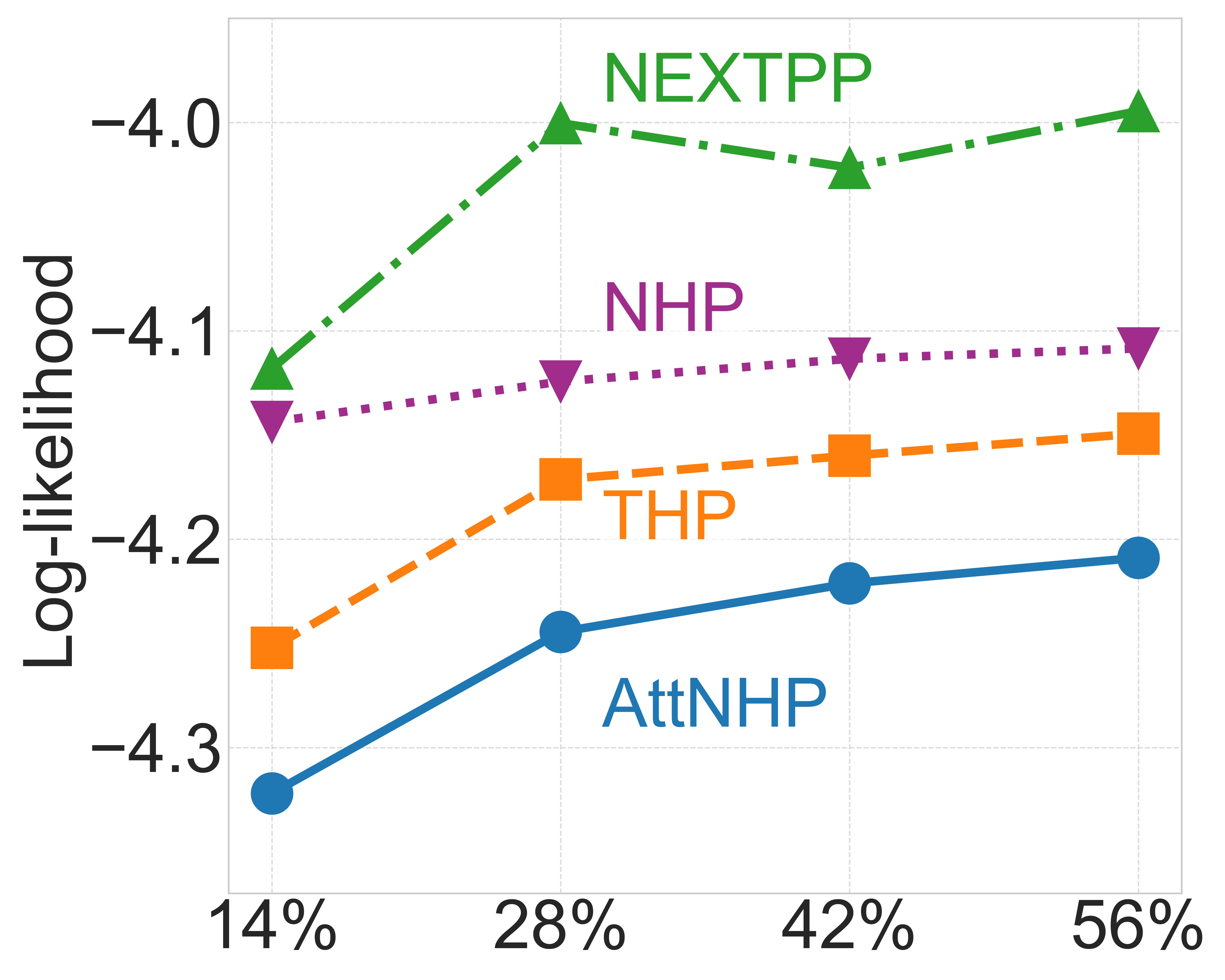}
}
\hfill
\subfloat[ErrRt(Retweet)]{%
    \includegraphics[width=0.233\linewidth]{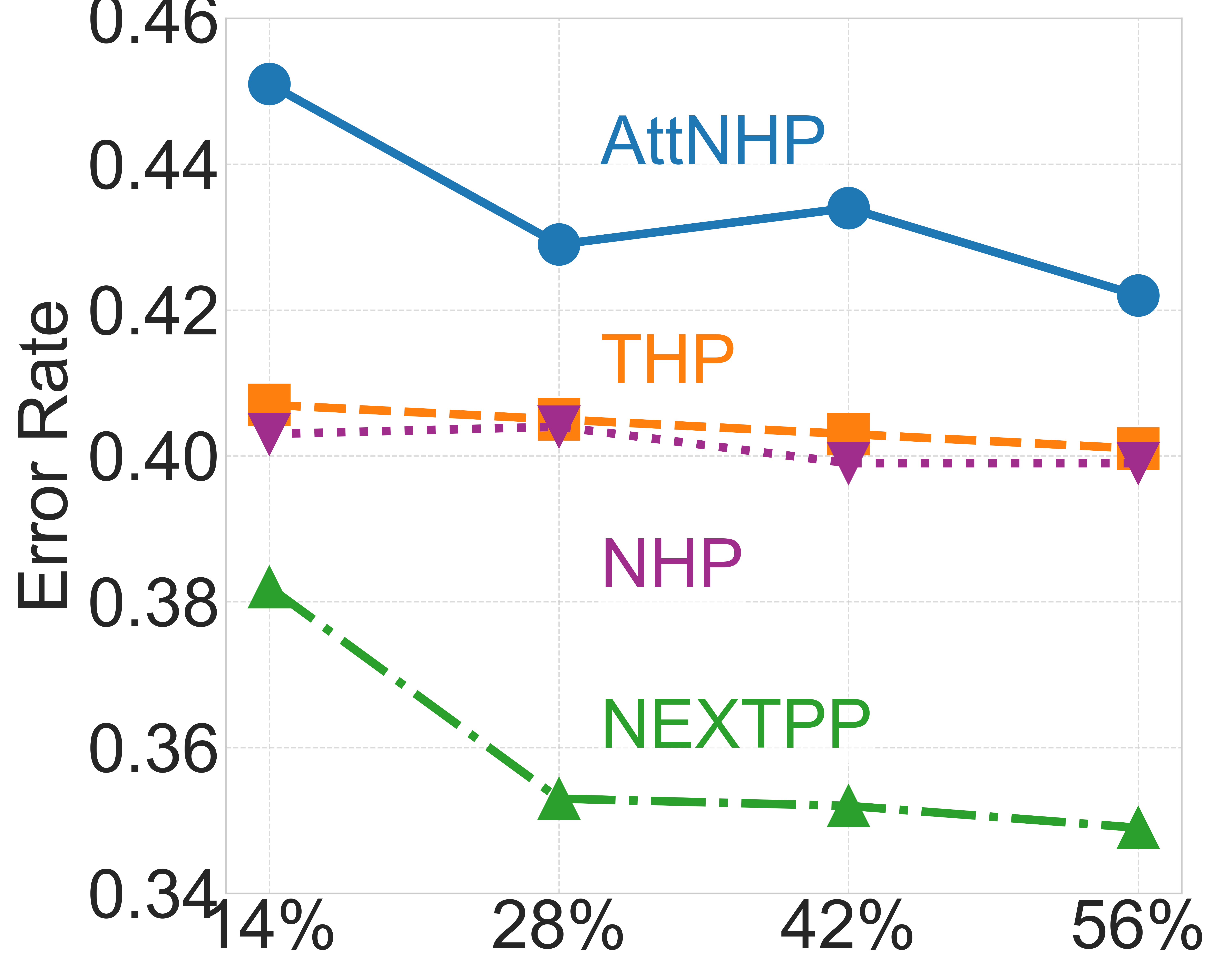}
}
\caption{Training curves with varying event counts (Percentage of the total training dataset) for Earthquake and Retweet datasets.}
\label{ER_num}
\end{figure}

\subsection{Effectiveness of $\mathbf{X}\text{-}$Interaction}
To demonstrate NEXTPP's ability to capture the influence of event history, we conduct comparative visualizations between the continuous-discrete interaction and conventional self-attention mechanisms. We visualize the same sequence showcased in Figure \ref{intro}, and the red vertical rectangle marks the occurrence time of the mainshock at 11.56 seconds, and display the dynamic evolution of heatmaps across several training epochs in Figure \ref{ca1}. Two key observations emerge. First, NEXTPP's continuous cross-attention rapidly concentrates its weight mass around the true preceding events, effectively tracing the causal impact of specific historical timestamps on each the evolved state representation. In contrast, self-attention remains broadly diffused, assigning non-negligible weight to distant or spurious positions and failing to identify the most important content and its impact. Second, as training progresses, NEXTPP’s attention heatmap develops clear, task-aligned concentration patterns that both sharpens and extends across variable temporal lags, demonstrating its capacity to adaptively attend to relevant intervals regardless of their distance. This pattern confirms that the cross-attention module endows NEXTPP with precise, history‐conditioned feature integration, crucial for intensity function support and timestamp forecasting in marked temporal point processes.

\subsection{Performance under Limited Training Data}
Figure \ref{ER_num} evaluates model performance on the Earthquake and Retweet datasets as the fraction of training events increases from 14\% to 56\%. Panels (a) and (c) plot average log-likelihood, while (b) and (d) show error rate. Across both tasks, NEXTPP (\textcolor{green}{green}) consistently achieves the highest log-likelihood and the lowest error rate at all data scales, demonstrating its ability to learn both accurate densities and precise event predictions. In contrast, Transformer-based models (THP, AttNHP) struggle when data are scarce and their performance plateaus or even slightly degrades as more events become available, likely because their attention heads latch onto spurious dependencies \cite{zuo2020transformer}. The Neural Hawkes Process (NHP) improves modestly with more data but remains outperformed by NEXTPP. These results confirm that NEXTPP's dual-channel cross-interaction mechanism robustly leverages both discrete and continuous dynamics, yielding superior and steadily improving performance as training volume grows.

\begin{figure}[t]
	\centering
	\subfloat[$ \tau $ {\footnotesize on different datasets}]{
         \label{sigma:tau}
		\includegraphics[scale=0.25]{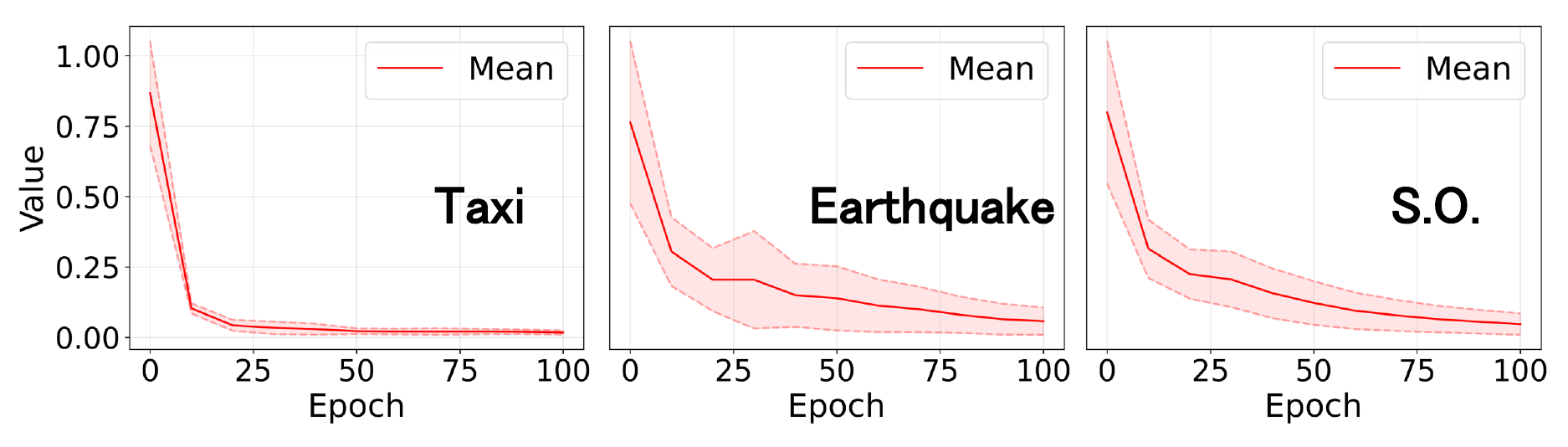}
	}%
    \subfloat[Training time(minutes)]{	
        \label{sigma:time}
		\includegraphics[scale=0.15]{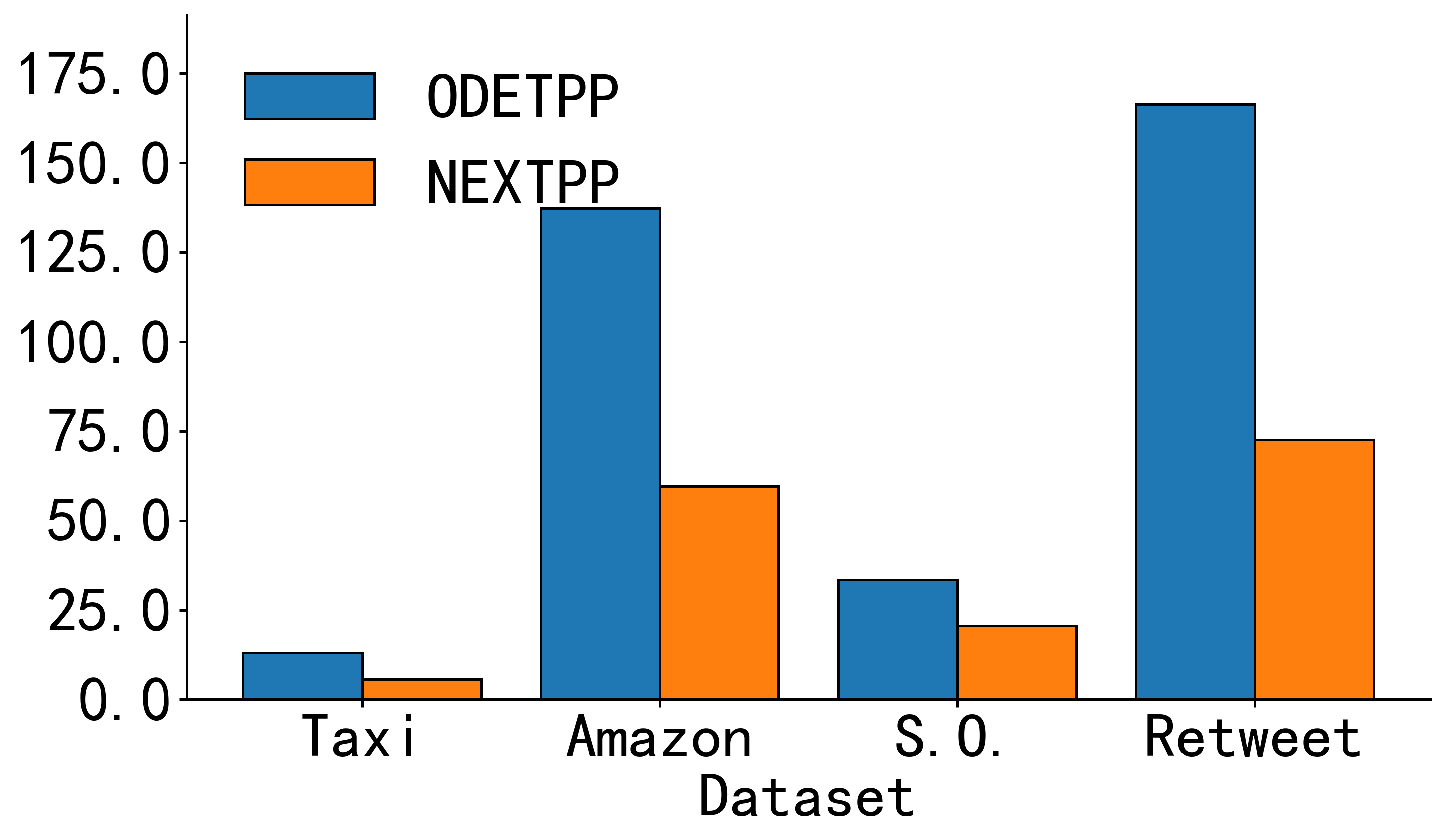}
	}%
	\label{sigma}
     \caption{Training Stability and Efficiency.}
\end{figure}
\subsection{Training Stability and Efficiency}
Figure \ref{sigma:tau} tracks the evolution of the latent space variance $\tau$ over 100 training epochs. $\tau$ rapidly decreases from initialization and converges to a narrow range with small cross-run variance, showing the trained Neural ODE parameters $ \theta $ produce well-conditioned dynamics $ f_\theta $ that stabilize the latent state space and avoid large corrections during training. Figure \ref{sigma:time} compares end-to-end training times on an NVIDIA RTX 4090 GPU between NEXTPP and the prior ODETPP model (both with 100 epochs and a single hidden layer of size 32). NEXTPP takes less than half the computation time, demonstrating its low-dimensional latent representation delivers major efficiency gains without losing stability.

\section{Conclusion}
In this work, we present NEXTPP, which successfully bridges the gap between discrete mark modeling and continuous-time dynamics in marked temporal point processes. By embedding event marks via self‑attention and evolving latent states through a Neural ODE, then fusing these representations with cross‑attention, NEXTPP explicitly captures how past marks influence future timing. Empirical evaluations across five diverse real‑world datasets demonstrate that NEXTPP consistently outperforms state‑of‑the‑art baselines in log‑likelihood and prediction error, while its cross‑attention weights offer insights into the bidirectional interplay of marks and timing.
\\

\textbf{Acknowledgements. }
This work is supported by National Natural Science Foundation of China (U22B2061), National Key R\&D Program of China (2022YFB4300603), and Natural Science Foundation of Sichuan, China (2024NSFSC0496).

\textbf{Disclosure of Interests. }
No competing interests exist.
\bibliography{references}

@article{whong2014foiling,
  title={FOILing NYC's taxi trip data},
  author={Whong, C.},
  journal={FOILing NYCs Taxi Trip Data},
  volume={18},
  pages={14},
  year={2014}
}

@misc{NiDataset2018,
  title={Amazon Product Review Dataset},
  author={Ni, J.},
  howpublished={\url{https://nijianmo.github.io/amazon/}},
  year={2018}
}

@misc{jure2014snap,
  title={SNAP Datasets: Stanford large network dataset collection},
  author={Jure, L.},
  howpublished={Retrieved December 2021 from \url{http://snap.stanford.edu/data}},
  year={2014}
}

@inproceedings{xue2023easytpp,
  title={EasyTPP: Towards Open Benchmarking Temporal Point Processes},
  author={Xue, Siqiao and Shi, Xiaoming and Chu, Zhixuan and Wang, Yan and Hao, Hongyan and Zhou, Fan and JIANG, Caigao and Pan, Chen and Zhang, James Y and Wen, Qingsong and others},
  booktitle={The Twelfth International Conference on Learning Representations},
  year={2023}
}

@inproceedings{zhou2013learning,
  title={Learning triggering kernels for multi-dimensional hawkes processes},
  author={Zhou, Ke and Zha, Hongyuan and Song, Le},
  booktitle={International conference on machine learning},
  pages={1301--1309},
  year={2013},
  organization={PMLR}
}

@inproceedings{du2016recurrent,
  title={Recurrent marked temporal point processes: Embedding event history to vector},
  author={Du, Nan and Dai, Hanjun and Trivedi, Rakshit and Upadhyay, Utkarsh and Gomez-Rodriguez, Manuel and Song, Le},
  booktitle={Proceedings of the 22nd ACM SIGKDD international conference on knowledge discovery and data mining},
  pages={1555--1564},
  year={2016}
}

@article{kidger2020neural,
  title={Neural controlled differential equations for irregular time series},
  author={Kidger, P. and Morrill, J. and Foster, J. and Lyons, T.},
  journal={Advances in Neural Information Processing Systems},
  volume={33},
  pages={6696--6707},
  year={2020}
}

@article{dupont2019augmented,
  title={Augmented neural odes},
  author={Dupont, Emilien and Doucet, Arnaud and Teh, Yee Whye},
  journal={Advances in neural information processing systems},
  volume={32},
  year={2019}
}

@article{song2020score,
  title={Score-based generative modeling through stochastic differential equations},
  author={Song, Yang and Sohl-Dickstein, Jascha and Kingma, Diederik P and Kumar, Abhishek and Ermon, Stefano and Poole, Ben},
  journal={arXiv preprint arXiv:2011.13456},
  year={2020}
}

@inproceedings{rodriguez2022lyanet,
  title={Lyanet: A lyapunov framework for training neural odes},
  author={Rodriguez, Ivan Dario Jimenez and Ames, Aaron and Yue, Yisong},
  booktitle={International conference on machine learning},
  pages={18687--18703},
  year={2022},
  organization={PMLR}
}

@article{mei2017neural,
  title={The neural hawkes process: A neurally self-modulating multivariate point process},
  author={Mei, Hongyuan and Eisner, Jason M},
  journal={Advances in neural information processing systems},
  volume={30},
  year={2017}
}

@inproceedings{zuo2020transformer,
  title={Transformer hawkes process},
  author={Zuo, Simiao and Jiang, Haoming and Li, Zichong and Zhao, Tuo and Zha, Hongyuan},
  booktitle={International conference on machine learning},
  pages={11692--11702},
  year={2020},
  organization={PMLR}
}

@inproceedings{zhang2020self,
  title={Self-attentive Hawkes process},
  author={Zhang, Qiang and Lipani, Aldo and Kirnap, Omer and Yilmaz, Emine},
  booktitle={International conference on machine learning},
  pages={11183--11193},
  year={2020},
  organization={PMLR}
}

@article{chen2020neural,
  title={Neural spatio-temporal point processes},
  author={Chen, Ricky TQ and Amos, Brandon and Nickel, Maximilian},
  journal={arXiv preprint arXiv:2011.04583},
  year={2020}
}

@inproceedings{yang2022transformer,
  title={Transformer Embeddings of Irregularly Spaced Events and Their Participants},
  author={Yang, Chenghao and Mei, Hongyuan and Eisner, Jason},
  booktitle={Proceedings of the Tenth International Conference on Learning Representations (ICLR)},
  year={2022}
}

@article{zhou2025non,
  title={Non-autoregressive diffusion-based temporal point processes for continuous-time long-term event prediction},
  author={Zhou, Wang-Tao and Kang, Zhao and Tian, Ling and Zhang, Jinchuan and Liu, Yumeng},
  journal={Expert Systems with Applications},
  volume={267},
  pages={126210},
  year={2025},
  publisher={Elsevier}
}

@article{rubanova2019latent,
  title={Latent ordinary differential equations for irregularly-sampled time series},
  author={Rubanova, Yulia and Chen, Ricky TQ and Duvenaud, David K},
  journal={Advances in neural information processing systems},
  volume={32},
  year={2019}
}

@book{kingman1992poisson,
  title={Poisson processes},
  author={Kingman, John Frank Charles},
  volume={3},
  year={1992},
  publisher={Clarendon Press}
}

@article{hawkes1971spectra,
  title={Spectra of some self-exciting and mutually exciting point processes},
  author={Hawkes, Alan G},
  journal={Biometrika},
  volume={58},
  number={1},
  pages={83--90},
  year={1971},
  publisher={Oxford University Press}
}

@inproceedings{zhou2022neural,
  title={Neural point process for learning spatiotemporal event dynamics},
  author={Zhou, Zihao and Yang, Xingyi and Rossi, Ryan and Zhao, Handong and Yu, Rose},
  booktitle={Learning for Dynamics and Control Conference},
  pages={777--789},
  year={2022},
  organization={PMLR}
}

@article{lewis1979simulation,
  title={Simulation of nonhomogeneous Poisson processes by thinning},
  author={Lewis, PA W and Shedler, Gerald S},
  journal={Naval research logistics quarterly},
  volume={26},
  number={3},
  pages={403--413},
  year={1979},
  publisher={Wiley Online Library}
}

@article{chen2018neural,
  title={Neural ordinary differential equations},
  author={Chen, Ricky TQ and Rubanova, Yulia and Bettencourt, Jesse and Duvenaud, David K},
  journal={Advances in neural information processing systems},
  volume={31},
  year={2018}
}

@inproceedings{zhang2024neural,
  title={Neural jump-diffusion temporal point processes},
  author={Zhang, Shuai and Zhou, Chuan and Liu, Yang Aron and Zhang, Peng and Lin, Xixun and Ma, Zhi-Ming},
  booktitle={Forty-first International Conference on Machine Learning},
  year={2024}
}

@article{upadhyay2018deep,
  title={Deep reinforcement learning of marked temporal point processes},
  author={Upadhyay, Utkarsh and De, Abir and Gomez Rodriguez, Manuel},
  journal={Advances in neural information processing systems},
  volume={31},
  year={2018}
}

@inproceedings{yuan2023spatio,
  title={Spatio-temporal diffusion point processes},
  author={Yuan, Yuan and Ding, Jingtao and Shao, Chenyang and Jin, Depeng and Li, Yong},
  booktitle={Proceedings of the 29th ACM SIGKDD Conference on Knowledge Discovery and Data Mining},
  pages={3173--3184},
  year={2023}
}

@article{ho2020denoising,
  title={Denoising diffusion probabilistic models},
  author={Ho, Jonathan and Jain, Ajay and Abbeel, Pieter},
  journal={Advances in neural information processing systems},
  volume={33},
  pages={6840--6851},
  year={2020}
}

@inproceedings{li2023smurf,
  title={SMURF-THP: score matching-based uncertainty quantification for transformer hawkes process},
  author={Li, Zichong and Xu, Yanbo and Zuo, Simiao and Jiang, Haoming and Zhang, Chao and Zhao, Tuo and Zha, Hongyuan},
  booktitle={International Conference on Machine Learning},
  pages={20210--20220},
  year={2023},
  organization={PMLR}
}

@article{isham1979self,
  title={A self-correcting point process},
  author={Isham, Valerie and Westcott, Mark},
  journal={Stochastic processes and their applications},
  volume={8},
  number={3},
  pages={335--347},
  year={1979},
  publisher={Elsevier}
}

@article{fox2016spatially,
  title={Spatially inhomogeneous background rate estimators and uncertainty quantification for nonparametric Hawkes point process models of earthquake occurrences},
  author={Fox, Eric Warren and Schoenberg, Frederic Paik and Gordon, Joshua Seth},
  journal={The Annals of Applied Statistics},
  volume={10},
  number={3},
  pages={1725},
  year={2016},
  publisher={Institute of Mathematical Statistics}
}

@inproceedings{sharma2021identifying,
  title={Identifying coordinated accounts on social media through hidden influence and group behaviours},
  author={Sharma, Karishma and Zhang, Yizhou and Ferrara, Emilio and Liu, Yan},
  booktitle={Proceedings of the 27th ACM SIGKDD conference on knowledge discovery \& data mining},
  pages={1441--1451},
  year={2021}
}

@article{xu2016patient,
  title={Patient flow prediction via discriminative learning of mutually-correcting processes},
  author={Xu, Hongteng and Wu, Weichang and Nemati, Shamim and Zha, Hongyuan},
  journal={IEEE transactions on Knowledge and Data Engineering},
  volume={29},
  number={1},
  pages={157--171},
  year={2016},
  publisher={IEEE}
}

@book{daley2003introduction,
  title={An introduction to the theory of point processes: volume I: elementary theory and methods},
  author={Daley, Daryl J and Vere-Jones, David},
  year={2003},
  publisher={Springer}
}

@article{blei2017variational,
  title={Variational inference: A review for statisticians},
  author={Blei, David M and Kucukelbir, Alp and McAuliffe, Jon D},
  journal={Journal of the American statistical Association},
  volume={112},
  number={518},
  pages={859--877},
  year={2017},
  publisher={Taylor \& Francis}
}

@inproceedings{higgins2017beta,
  title={beta-vae: Learning basic visual concepts with a constrained variational framework},
  author={Higgins, Irina and Matthey, Loic and Pal, Arka and Burgess, Christopher and Glorot, Xavier and Botvinick, Matthew and Mohamed, Shakir and Lerchner, Alexander},
  booktitle={International conference on learning representations},
  year={2017}
}

\end{document}